\documentclass[final]{cvpr}

\usepackage{times}
\usepackage{epsfig}
\usepackage{graphicx}
\usepackage{amsmath}
\usepackage{amssymb}

\usepackage{booktabs}       
\usepackage[symbol]{footmisc}
\usepackage{xr-hyper}
\usepackage{array}
\usepackage{amsthm}
\usepackage{amsfonts}
\usepackage{mathtools}
\usepackage{algorithm}
\usepackage{algorithmic}
\usepackage{subfig}
\usepackage[export]{adjustbox}
\usepackage{verbatim}
\newtheorem{theorem}{Theorem}[]

\newtheorem{lemma}[theorem]{Lemma}
\newtheorem{proposition}[theorem]{Proposition}
\mathchardef\mhyphen="2D
\usepackage{xr}

\usepackage[pagebackref=true,breaklinks=true,colorlinks,bookmarks=false]{hyperref}

\def\cvprPaperID{****} 
\def\confYear{ICCV 2021}

\begin{document}

\title{Dual Projection Generative Adversarial Networks for Conditional Image Generation}

\author{
Ligong Han,\textsuperscript{\rm 1}
Martin Renqiang Min,\textsuperscript{\rm 2}
Anastasis Stathopoulos,\textsuperscript{\rm 1}\\
Yu Tian,\textsuperscript{\rm 1}
Ruijiang Gao,\textsuperscript{\rm 3}
Asim Kadav,\textsuperscript{\rm 2}
Dimitris Metaxas\textsuperscript{\rm 1}\\
\textsuperscript{\rm 1}Department of Computer Science, Rutgers University~
\textsuperscript{\rm 2}NEC Labs America\\
\textsuperscript{\rm 3}McCombs School of Business, The University of Texas at Austin\\
lh599@rutgers.edu, ~renqiang@nec-labs.com, ~as2947@cs.rutgers.edu\\ yt219@cs.rutgers.edu, ~ruijiang@utexas.edu, ~asim@nec-labs.com, ~dnm@cs.rutgers.edu
}

\maketitle

\begin{abstract}
   Conditional Generative Adversarial Networks (cGANs) extend the standard unconditional GAN framework to learning joint data-label distributions from samples, and have been established as powerful generative models capable of generating high-fidelity imagery. A challenge of training such a model lies in properly infusing class information into its generator and discriminator. For the discriminator, class conditioning can be achieved by either (1) directly incorporating labels as input or (2) involving labels in an auxiliary classification loss. In this paper, we show that the former directly aligns the class-conditioned fake-and-real data distributions $P(\text{image}|\text{class})$ ({\em data matching}), while the latter aligns data-conditioned class distributions $P(\text{class}|\text{image})$ ({\em label matching}). Although class separability does not directly translate to sample quality and becomes a burden if classification itself is intrinsically difficult, the discriminator cannot provide useful guidance for the generator if features of distinct classes are mapped to the same point and thus become inseparable. Motivated by this intuition, we propose a Dual Projection GAN (P2GAN) model that learns to balance between {\em data matching} and {\em label matching}. We then propose an improved cGAN model with Auxiliary Classification that directly aligns the fake and real conditionals $P(\text{class}|\text{image})$ by minimizing their $f\mhyphen\text{divergence}$. Experiments on a synthetic Mixture of Gaussian (MoG) dataset and a variety of real-world datasets including CIFAR100, ImageNet, and VGGFace2 demonstrate the efficacy of our proposed models.
\end{abstract}

\section{Introduction}
\begin{figure*}[h]
    \centering
    \includegraphics[width=1\linewidth]{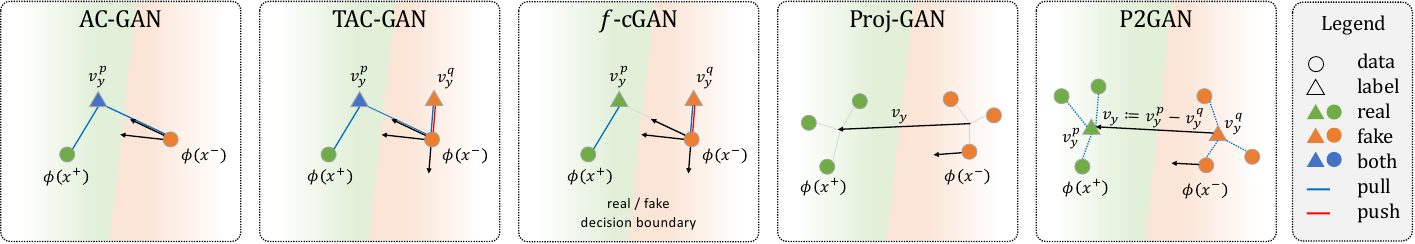}
    \caption{Illustrative figures visualize the learning schema of conditional discriminator losses. The color indicates how real/fake embeddings interact during {\em discriminator training}. The green/red boundary indicates (unconditional) real/fake decision boundary. Triangles represent class embeddings, and circles indicate image embeddings (\eg $v_y^p$ is blue triangle in AC-GAN since it is class embedding and is trained on both real and fake data). Solid blue and red lines represent pull/push forces respectively. Black arrows indicate the forces that a fake image embedding receives.}
    \label{fig:intuition}
\end{figure*}
Generative Adversarial Networks (GANs)~\cite{goodfellow2014generative} are an algorithmic framework that allows implicit generative modeling of data distribution from samples~\cite{mohamed2016learning}. It has attracted great attention due to its ability to model very high-dimensional data, such as images or videos and to produce sharp and faithful samples~\cite{karras2017progressive,zhang2018self,brock2018large}. Conditional GAN (cGAN)~\cite{mirza2014conditional,odena2017conditional,miyato2018cgans} is an extension of GAN that utilizes the label information and aims to learn the joint distribution of data and label. Thanks to its ability to control over the generative process by conditioning on labels, it has been widely adopted in real-world problems including class-conditional image generation~\cite{nguyen2017plug,odena2017conditional,choi2018stargan}, text-to-image generation~\cite{reed2016generative,Zhang2016StackGANTT}, image-to-image generation~\cite{isola2017image,zhu2017unpaired}, text-to-video synthesis~\cite{text2video,tfgan,han2018learning,han2020robust}, domain adaptation~\cite{hoffman2018cycada,mitash2020scene} \etc

Different conditional GANs differ in the way how data and labels are incorporated in the discriminator. Class conditioning can be achieved by either (1) conditioning the discriminator directly on labels or their embeddings~\cite{mirza2014conditional,isola2017image,miyato2018cgans}, or by (2) incorporating an auxiliary classification loss in the objective, as in (T)AC-GANs\cite{mirza2014conditional,gong2019twin}. Recently, cGAN has received a major update that changed the label from being concatenated~\cite{mirza2014conditional,reed2016generative,Zhang2016StackGANTT} to being projected~\cite{miyato2018cgans}. The projection discriminator takes the inner product between the label embedding and the data/image embedding, and remains to be the choice of many state-of-the-art methods~\cite{zhang2018self,brock2018large}.

In this paper, we first give insights on projection discriminators. We point out that the success of projection can be explained by its flexible form. By tying real and fake class embeddings as their difference, a projection discriminator can ideally realize two extremes in a single form: (1) matching conditional label distributions $P(\text{class}|\text{image})$ ({\em label matching}), and (2) matching conditional data distributions $P(\text{image}|\text{class})$ ({\em data matching}). Moreover, the visualization of data embeddings of trained projection discriminators does not show obvious patterns of class clustering. This suggests that projection may bias towards conditional data matching. Although label matching does not directly translate to the fidelity of generated samples, it is still desirable for generating high-quality images. For example, the discriminator would not provide useful guidance for the generator if features of distinct classes are mapped to the same point and thus become inseparable.

To this end, we propose a new conditional generative adversarial network, namely {\em Dual Projection GAN} (P2GAN). The main feature of our design is to inherit the flexibility of projection while performing explicit label matching. Realized by auxiliary classification losses, label matching enables the discriminator to exploit useful information for the generator. However, if such task is intrinsically difficult (such as ImageNet~\cite{russakovsky2015imagenet}), label matching becomes a burden. In an extreme case when the auxiliary classification task failed completely, (T)AC-GANs will degrade to an unconditional GAN while P2GAN degrades to a projection GAN. We also present adaptive approaches to weighing data matching and label matching. Furthermore, we respectively propose two variants for explicit data matching and label matching: (1) direct data matching GAN (DM-GAN), (2) and $f\mhyphen\text{cGAN}$ which aligns the fake and real conditionals $P(\text{class}|\text{image})$ by minimizing their $f\mhyphen\text{divergence}$. Finally, we conduct extensive experiments on various datasets to show the efficacy of the proposed models.

\section{Related Work}
\noindent \textbf{Projection.}
The key idea of Proj-GAN is to tie parameters of AC and twin AC. This design brings flexibility in modeling data matching and label matching. We leverage its flexible design in our P2GAN model. The same projection form is considered in projection-based MINE~\cite{belghazi2018mine,han2020unbiased}.

\noindent \textbf{Multi-task learning.}
Training GANs with both discrimination loss and classification loss can be viewed as a multi-task learning problem. \cite{kendall2018multi} propose a principled approach~\cite{kendall2018multi} that weighs multiple loss functions by considering the homoscedastic uncertainty of each task. The proposed P2GAN-w model is built upon this notion.

\noindent \textbf{$f\mhyphen\text{divergence}$ in GANs.}
\cite{nowozin2016f} formulated a family of GANs into a variational $f\mhyphen\text{divergence}$ minimization framework~\cite{nowozin2016f}. Their proposed $f\mhyphen\text{GAN}$ is for marginal matching (unconditional GAN) while our $f\mhyphen\text{cGAN}$ aims to minimize the $f\mhyphen\text{divergence}$ between real and fake conditionals and does not optimize its dual form.

\noindent \textbf{Twin auxiliary classifiers.}
TAC-GAN corrects the bias in AC-GANs by introducing a twin AC and maximizing its classification loss on generated samples. A binary version of twin AC has been introduced as an Anti-Labeler in CausalGAN~\cite{kocaoglu2017causalgan}. CausalGAN has observed unstable training behavior after introducing such binary twin ACs. Rob-GAN~\cite{liu2019rob} splits the categorical classification losses of real and fake data in AC-GAN, however, it does not have a twin AC (dual projections) and does not aim to balance label-matching and data-matching.

\begin{figure*}[h]
    \centering
    \includegraphics[width=0.95\linewidth]{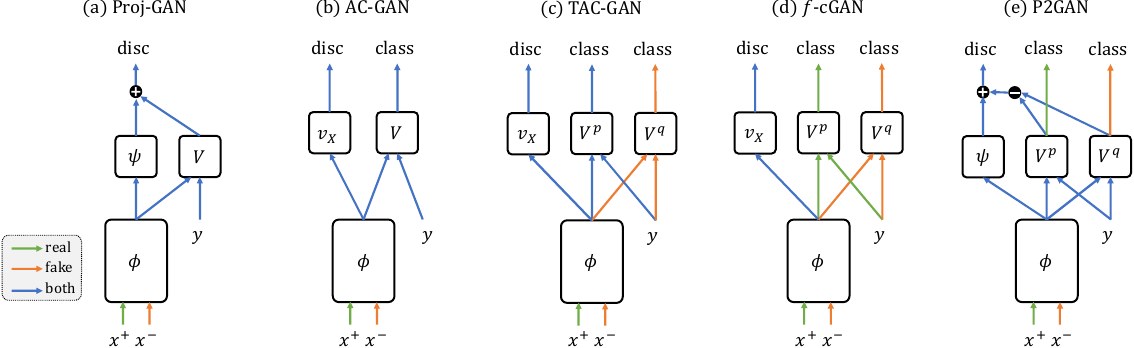}
    \caption{Discriminator models for conditional GANs. Solid green lines represent the flow path of real data, red line represent the flow path of generated data and blue lines mean both real and generated data flow through them. $x^+$ represents real images, ${x}^-$ are generated images and $y$ are labels. Matrix $V^p$ and $V^q$ are the collection of class embeddings $\{v^p_y\}$ and $\{v^q_y\}$ for twin auxiliary classifiers, respectively.}
    \label{fig:flowchart}
\end{figure*}

\section{Method}
\subsection{Background}
Throughout the paper, we denote data-label pairs as $\{x_i, y_i\}^n_{i=1} \subseteq \mathcal{X}\times\mathcal{Y}$ drawn from the joint distribution $P_{XY}$, where $x$ is an image (or other forms of data) and $y$ is its label, \ie, $\mathcal{Y}=\{1, \ldots, K\}$. A generator is trained to transform samples $z \sim P_Z$ from a canonical distribution conditioned on labels to match the real data distributions. {\em Real} distributions are denoted as $P$ and {\em generated} distributions are denoted as $Q$. We also denote real and fake data as $x^+ \sim P_X$ and $x^- \sim Q_X$, respectively. The discriminator of a conditional GAN learns to distinguish samples drawn from the joint distribution $P_{XY}$ and $Q_{XY}$. Its objectives can be written as:
\begin{align}
    L_{D} =& \mathbb{E}_{x,y \sim P_{XY}}{\mathcal{A}(-\tilde{D}(x,y))} + \label{eq:cgan}\\
    &\mathbb{E}_{z \sim P_Z, y \sim Q_Y}{\mathcal{A}(\tilde{D}(G(z,y),y))}, \quad \text{and} \nonumber \\
    L_{G} =& \mathbb{E}_{z \sim P_Z, y \sim Q_Y}{\mathcal{A}(-\tilde{D}(G(z,y),y))}. \label{eq:cgan_g}
\end{align}
\noindent Here $\mathcal{A}$ is the {\em activation function} and $\tilde{D}$ is the {\em logit} or discriminator's output before activation. Note that choosing $\mathcal{A}(t)={softplus}(t)=\log{(1+\exp{(t)})}$ recovers the original GAN formulation\cite{goodfellow2014generative}. With this activation function, 
the logit of an optimal discriminator in Equation~\ref{eq:cgan} can be decomposed in two ways,
\begin{align}
    \tilde{D}^*(x,y) &= \underbrace{\log\frac{P(x)}{Q(x)}}_{\text{Marginal Matching}} + \underbrace{\log\frac{P(y|x)}{Q(y|x)}}_{\text{Label Matching}} \label{eq:decomp1}\\
    &= \underbrace{\log\frac{P(x|y)}{Q(x|y)}}_{\text{Data Matching}} + \log\frac{P(y)}{Q(y)}.
    \label{eq:decomp2}
\end{align}
\noindent From Equation~\ref{eq:decomp1}, one can derive the logit of a {\em projection} discriminator~\cite{miyato2018cgans},
\begin{align}
    \tilde{D}(x,y) = v_y^\text{T}\phi(x) + \psi(\phi(x)),
    \label{eq:proj}
\end{align}
\noindent where $\phi(\cdot)$ is the image embedding function, $v_y$ is embedding of class $y$, and $\psi$ collects the residual terms. $v_y \vcentcolon = v^p_y - v^q_y$ is the difference of real and fake class embeddings.

\subsection{Projection and Dual Projection GANs}
\noindent \textbf{Effectiveness of tying class embeddings.}
The key idea of a projection discriminator~\cite{miyato2018cgans} is to tie the parameters $v^p_y$ and $v^q_y$ into a single $v_y$. Intuitively, tying embeddings allows a projection discriminator to turn the problem of learning categorical decision boundaries to learning a relative translation vector for each class. The latter is much easier than the former. Intuitively, it is in general easier to learn a relative comparison than to learn each parameter in an absolute scale. For example, learning $v^p_y$ is a hard problem by itself when training on large scale datasets like ImageNet~\cite{russakovsky2015imagenet}.

Without loss of generality, let's assume $\psi(\cdot)$ as a linear function $v_\psi$. From Equation~\ref{eq:cgan} we can see that $L_{D}$ is trying to maximize $(v_y+v_\psi)^\text{T}\phi(x^+)$ and to minimize $(v_y+v_\psi)^\text{T}\phi(x^-)$. If we approximate \textit{softplus} function by $ReLU=\max{(0, \cdot)}$,
we observe that the loss is large when $x^+$ and $x^-$ are mis-classified, under which $L_D \approx (v_y+v_\psi)^\text{T}(\phi(x^-)-\phi(x^+))$. Intuitively, the learning procedure of Proj-GAN can be understood as performing two alternative steps,
\begin{align}
    &\text{D-step}: ~\text{Align}~ (v_y+v_\psi) ~\text{with}~ (\phi(x^+)-\phi(x^-)); \nonumber \\
    &\text{G-step}: ~\text{Move}~ \phi(x^-) ~\text{along}~ (v_y+v_\psi). \nonumber
\end{align}
This shows that by tying parameters, Proj-GAN is able to directly perform {\em data matching} (align $Q(x|y)$ with $P(x|y)$) without explicitly enforcing {\em label matching}. 

\noindent \textbf{Enforcing label matching.}
Ideally, $v_y$ should recover the difference between underlying $v^p_y$ and $v^q_y$, but there is no explicit constraint to enforce such property. To do this, we start by untying the class embeddings $v_y \vcentcolon= v^p_y - v^q_y$. Then, we encourage $V^p$ and $V^q$ to learn conditional distributions ${p(y|x)}$ and ${q(y|x)}$, respectively\footnote{Matrix $V$ is the collection of class embeddings $v_y$ for all categories.}. This is usually done by minimizing cross-entropy losses with a $softmax$ function,
\begin{align}
    L^{p}_{mi} &= -v_y^{p\text{T}}\phi(x^+)+\log\sum_{y'}{\exp({v_{y'}^{p\text{T}}\phi(x^+)})}, ~\text{and} \nonumber \\
    L^{q}_{mi} &= -v_y^{q\text{T}}\phi(x^-)+\log\sum_{y'}{\exp({v_{y'}^{q\text{T}}\phi(x^-)})}. \label{eq:t_ce}
\end{align}
Note that the two classifiers $V^p$ and $V^q$ are trained on real and fake data respectively. This is different from AC-GAN and TAC-GAN where the (first) discriminator is trained on both $x^+$ and $x^-$. This is shown in Figure~\ref{fig:intuition} and ~\ref{fig:flowchart}.

\noindent \textbf{Dual Projection GAN (P2GAN).} The discriminator and generator losses for P2GAN is given as follows,
\begin{align}
    L_{D}^{P2} &= L_D(\tilde{D}) + L^{p}_{mi} + L^{q}_{mi},  \\
    \text{and} ~~ L_{G}^{P2} &= L_G(\tilde{D}),  \qquad  \nonumber\\
    \text{with} \quad~ \tilde{D} &= (v_y^p-v_y^q)^\text{T}\phi(x) + \psi(\phi(x)) \nonumber
\end{align}
\noindent The proposed method has untied projection vectors, thus we term the model Dual Projection GAN, or P2GAN. Note that both $L_D(\tilde{D})$ and $L^{p}_{mi}$ contain parameter $V^p$, and similarly, both $L_D(\tilde{D})$ and $L^{q}_{mi}$ contain parameter $V^q$. It is also worth mentioning that similar to the Proj-GAN, P2GAN only proposes a form of the logit while leaves the activation function free of choices.

We observe that when cross-entropy losses are removed, P2GAN is equivalent to Proj-GAN except that $v_y$ is over-parameterized as $v_y^p-v_y^q$. This suggests that when class separability is intrinsically difficult to model and the two cross-entropy losses are not effectively optimized, P2GAN reduces to a Proj-GAN. In this case, a model that relies on {\em marginal-conditional}~\cite{li2017alice} decomposition, such as an AC-GAN or a TAC-GAN, degrades to an unconditional GAN instead. This illustrates that the proposed P2GAN {\em implicitly} balances data matching and label matching by inheriting the Proj-GAN. We can also {\em explicitly} weigh $L^{p}_{mi}$ and $L^{q}_{mi}$.

\noindent \textbf{Weighted Dual Projection GAN (P2GAN-w).}
From a multi-task learning perspective~\cite{kendall2018multi}, it should be beneficial to let the model learn to weigh {\em data matching} and {\em label matching}. Here we consider adding a ``gate'' between the two losses,
\begin{align}
    L^{P2w}_{D} = (1-\lambda) \cdot L_{D} + \lambda \cdot (L^p_{mi}+L^q_{mi}).
    \label{eq:p2w}
\end{align}
We tested three variants of weighted P2GAN: (1) exponential decay, (2) scalar valued, and (3) amortised.

\noindent \textbf{P2GAN-d.} The simplest way is to define $\lambda$ as a decaying factor. We set $\lambda = e^{-t/T}$, $t$ is the number of iterations during training time.

\noindent \textbf{P2GAN-s.} Let $\lambda \in [0, 1]$ be a learnable parameter and initialized as $0.5$. When $\lambda=0$, P2GAN-s reduces to a Proj-GAN. When $\lambda>0$, class separation is explicitly enforced. The data matching task is intuitively much easier than the label matching task in the early stage of training and $\lambda$ would quickly vanish. To accommodate this, we follow~\cite{kendall2018multi} and add a penalty term on $\lambda$ (P2GAN-sp),
\begin{align}
    L^{P2sp}_{D} = &(1-\lambda) \cdot L_{D} + \lambda \cdot (L^p_{mi}+L^q_{mi}) \nonumber\\
    &- \frac{1}{2}\log{\lambda (1-\lambda)}.
    \label{eq:p2sp}
\end{align}

\noindent \textbf{P2GAN-a.} We also experimented with learning amortized homoscedastic weights for each data point. $\lambda(x) \ge 0$ is then a function of $x$ (P2GAN-a). Since $\lambda(x)$ are per-sample weights, in Equation~\ref{eq:p2w} it should be inside of the expectation. We defer its lengthy definition in Supplementary. Likewise, a penalty term can be added (P2GAN-ap). Also, when loss terms involve non-linearity in the mini-batch expectation, for example, the $\log$ in MINE~\cite{belghazi2018mine}, any type of ``linearization'' tricks can be applied~\cite{mroueh*2020improved}.

Comparisons of different weighting strategies are given in experiments Table~\ref{tab:p2w_all}. In practice we find $\lambda(x)$ with penalty works best. 
In following Experiments, P2GAN-w stands for P2GAN-ap, if not specified.

\subsection{Theoretical Analysis}
As discussed previously, a projection discriminator defined in Equation~\ref{eq:proj} can flexibly transit between two decomposition forms. Here we discuss two variants of conditional GAN models that perform {\em only} data matching {\em or} label matching.

\noindent \textbf{cGAN that performs data matching (DM-GAN).} Setting $\psi(\cdot)$ as zero\footnote{In practice spectral normalization~\cite{miyato2018spectral} is often applied and this null solution for $\psi$ cannot be reached.}, Proj-GAN can be viewed as a weighted sum of $K$ unconditional GAN objectives with binary cross-entropy losses (where $K$ is the number of classes):
\begin{proposition}
When $\psi=0$, a Proj-GAN reduces to $K$ unconditional GANs, each of them minimizes the Jensen-Shannon divergence between $P_{X|y}$ and $Q_{X|y}$ with mixing ratio $\{\frac{P(y)}{P(y)+Q(y)}, \frac{Q(y)}{P(y)+Q(y)}\}$. Its value function can be written as,
\begin{align}
    \mathbb{E}_{P_Y}\left\{ \mathbb{E}_{P_{X|Y}}\log{D(x|y)}+\frac{Q_y}{P_y}\mathbb{E}_{Q_{X|Y}}\log{(1-D(x|y))} \right\}.\nonumber
\end{align}
\label{prop1}
\end{proposition}

We observe a slight improvement in terms of FID score from our early experiment on CIFAR100\footnote{this conclusion is not reflected in Table~\ref{tab:ablation} due to different hyper-parameters and configurations.}, which might suggest that Proj-GAN biases towards data matching than label matching. DM-GAN is also the cGAN implementation in~\cite{Mescheder2018ICML,Karras2020ada}.

\noindent \textbf{cGAN that performs label matching ($f\mhyphen\text{cGAN}$).} At the other end of the spectrum, if we explicitly model the marginal matching and label matching terms in Equation~\ref{eq:decomp1}, we arrive at a (T)AC-GAN-like model which follows the {\em marginal-conditional} decomposition. To be precise, marginal matching can be achieved via (unconditional) GAN loss, and label matching can be enforced by minimizing their divergence.
\begin{proposition}
Given a generator $G$, if cross entropy losses $L^{p}_{mi}$ and $L^{q}_{mi}$ are minimized optimally, then the difference of two losses evaluated at fake data equals the reverse KL-divergence between $P_{Y|X}$ and $Q_{Y|X}$,
\begin{align}
    L^{p}_{mi}(x^-)-L^{q}_{mi}(x^-) = \mathbb{E}_{Q_{X}}{\text{KL}(Q_{Y|X} \Vert P_{Y|X})}.
\end{align}
\label{prop2}
\end{proposition}
From the above proposition we derive a new conditional GAN,
\begin{align}
    L_{D}^{f} &= L_D(\tilde{D}) + L^{p}_{mi}(x^+) + L^{q}_{mi}(x^-),  \nonumber\\
    \text{and} ~~ L_{G}^{f} &= L_G(\tilde{D}) + L^{p}_{mi}(x^-) - L^{p}_{mi}(x^-),  \nonumber\\
    \text{with} \quad \tilde{D} &= v_X^\text{T}\phi(x) + b \label{eq:fc}
\end{align}
In fact, $L^{p}_{mi}(x^-) - L^{p}_{mi}(x^-) = \mathbb{E}_{Q_{XY}}-\log \frac{Q^{p}(y|x)}{Q^{q}(y|x)}$. By replacing the $-\log$ with function $f$, we can also generalize the reverse-KL to $f\mhyphen\text{divergence}$~\cite{nowozin2016f}, $\mathbb{E}_{Q_{XY}}{f\left(\frac{P(y|x)}{Q(y|x)}\right)} = \mathbb{E}_{Q_{X}}D_{f}(P_{Y|X}\Vert Q_{Y|X})$. The generator loss then becomes
\begin{align}
    L_{G}^{f} &= L_G + \mathbb{E}_{Q_{XY}}{f(\exp(T^p(x,y) - T^q(x,y)))}.\label{eq:fc_g}
\end{align}
\noindent Here, $T^p = v_y^{p\text{T}}\phi(x)-\log\sum_{y'}{\exp({v_{y'}^{p\text{T}}\phi(x)})}$ is an estimator of $\log P(y|x)$, and similarly $T^q$ is an estimator of $\log Q(y|x)$. Equation~\ref{eq:fc} and~\ref{eq:fc_g} defines a general conditional GAN framework, which we call $f\mhyphen\text{cGAN}$.

To show that $f\mhyphen\text{cGAN}$ is theoretically sound and Nash equilibrium can be achieved, we provide the following theorem.
\setcounter{theorem}{0}
\begin{theorem}
Denoting $P_{XY}$ and $Q_{XY}$ as the data distribution and the distribution induced by $G$, their Jensen-Shannon divergence is upper bounded by the following,
\begin{align}
    & \text{JSD}(P_{XY},Q_{XY}) \le  \\
    & 2c_1 \sqrt{2 \text{JSD}(P_X, Q_X)} + c_2 \sqrt{2 \text{KL}(P_{Y|X} \Vert Q^{p}_{Y|X})} + \nonumber\\
    & c_2 \sqrt{2 \text{KL}(Q_{Y|X} \Vert Q^{q}_{Y|X})} + c_2 \sqrt{2 \text{KL}(Q^{q}_{Y|X} \Vert Q^{p}_{Y|X})}.\nonumber
\end{align}
\label{thm1}
\end{theorem}
\noindent in the above, constants $c_1$ and $c_2$ are upper bounds of $\frac{1}{2}\int{|P_{Y|X}(y|x)|\mu(x,y)}$ and $\int{|Q_{X}(x)|\mu(x)}$, respectively, where $\mu$ is a $\sigma\mhyphen\text{finite}$ measure.

\subsection{Comparison with Other Methods}
First, dual projection GAN extends a projection GAN by untying class embeddings and enforcing class separability in each domain. Comparing P2GAN with $f\mhyphen\text{cGAN}$, both of them minimizes $L^{p}_{mi}(x^+)+L^{q}_{mi}(x^-)$ in the discriminator loss. As for the generator loss, $f\mhyphen\text{cGAN}$ minimizes $L^{p}_{mi}(x^-)-L^{q}_{mi}(x^-)$, while P2GAN enforces label matching via $L_G((v_y^p-v_y^q)^\text{T}\phi(x^-) + \psi(\phi(x^-)))$ thus the $LogSumExp$ function is not involved.

By adding a term $L^p_{mi}(x^-)$ to $L^f_D$ in Equation~\ref{eq:fc}, one can recover the TAC-GAN\footnote{Note that this term is already omitted in its actual implementation, however, it is indispensable for the theoretical analysis to hold.}. It is worth noting that this term is required since TAC-GAN aims to chain $\mathbb{E}_{P_{X}}{\text{KL}(P_{Y|X} \Vert Q^{p}_{Y|X})}$ and $\mathbb{E}_{Q_{X}}{\text{KL}(Q_{Y|X} \Vert Q^{p}_{Y|X})}$ together, while our $f\mhyphen\text{cGAN}$ tries directly align $Q_{Y|X}$ with $P_{Y|X}$. In experiments we show that this simple trick can largely boost performance. Further, by removing all terms related to $L^q_{mi}$ we derive AC-GAN.

Both P2GAN and $f\mhyphen\text{cGAN}$ only model {\em data-to-class} relations, and {\em data-to-data} relation is indirectly modeled via class embeddings. While ContraGAN~\cite{kang2020contragan} and ReACGAN~\cite{kang2021rebooting} models these two relations directly. It is orthogonal to our method and we leave it for future work. 

\begin{table}[h]
    \caption{FID scores of different weighting strategies for P2GAN-w. All models are trained for 80000 iterations on ImageNet, 62000 iterations on CIFAR100, and 50000 iterations on VGGFace200.}
    \label{tab:p2w_all}
    \centering
    \resizebox{1\linewidth}{!}{
    \begin{tabular}{lccc}
    \toprule
    Datasets         & ImageNet & CIFAR100 & VGGFace200 \\
    \hline
    P2GAN            & 19.22  & 10.55 & 23.15 \\
    P2GAN-d (T=200)  & -      & 10.51 & 21.24 \\
    P2GAN-d (T=2000) & -      & 10.35 & 24.49 \\
    P2GAN-s          & 18.62   & {\bf 9.03} & 20.59 \\
    P2GAN-sp         & 20.68   & 9.51 & 20.18 \\
    P2GAN-a          & -      & 10.13 & 20.26 \\
    P2GAN-ap         & {\bf 18.31}   & 9.82 & {\bf 18.99} \\
    \bottomrule
    \end{tabular}}
\end{table}
\begin{table}[h]
    \caption{Average rank of Proj-GAN, TAC-GAN, $f\mhyphen\text{cGAN}$ and P2GAN on MoG dataset. For each method, the average ranks of using BCE loss, hinge loss, and all experiments are reported.}
    \label{tab:mog_rank}
    \centering
    \resizebox{1\linewidth}{!}{
    \begin{tabular}{lcccc}
    \toprule
    Models     & Proj-GAN & TAC-GAN* & $f\mhyphen\text{cGAN}$ (ours)   &  P2GAN (ours) \\
    \hline
    BCE   & 3.90 & 2.45 & 2.15 & {\bf 1.50} \\
    Hinge & {\bf 1.80} & 3.35 & 2.45 & 2.40 \\
    \hline
    Overall    & 2.85 & 2.90 & 2.30 & {\bf 1.95} \\
    \bottomrule
    \end{tabular}}
\end{table}
\begin{table}[h]
    \caption{Max-FID scores for 1D MOG synthetic dataset.}
    \vspace{-0.1in}
    \label{tab:mog_fid}
    \centering
    \resizebox{1\linewidth}{!}{
    \begin{tabular}{lccccc}
    \toprule
    {BCE~/~Hinge} & $d_m=1$ & $d_m=2$ & $d_m=3$ & $d_m=4$ & $d_m=5$ \\
    \hline
    Proj-GAN &  0.0059  &  0.0213  &  0.0291  &  0.0339  &  0.0796 \\
    TAC-GAN* &  {\bf 0.0029}  &  0.0090  &  0.0134  &  {\bf 0.0169}  &  0.0243 \\
    $f$-cGAN &  0.0032  &  {\bf 0.0072}  &  {\bf 0.0129}  &  0.0184  & {\bf 0.0213} \\
    P2GAN &  {\bf 0.0026}  &  {\bf 0.0060}  &  {\bf 0.0085}  &  {\bf 0.0124}  &  {\bf 0.0212} \\
    \hline
    Proj-GAN &  {\bf 0.0200}  &  {\bf 0.0481}  &  {\bf 0.0733 }  &  0.1253  &  0.2066 \\
    TAC-GAN* &  0.0270  &  0.0887  &  0.1307  &  0.1483  &  0.2285 \\
    $f$-cGAN &  {\bf 0.0236}  &  0.0829  &  0.1187  &  {\bf 0.0847}  &  {\bf 0.1100} \\
    P2GAN &  0.0245  &  {\bf 0.0473}  &  {\bf 0.0901}  &  {\bf 0.1076}  &  {\bf 0.1698} \\
    \bottomrule
    \end{tabular}}
    \vspace{-0.1in}
\end{table}
\begin{table*}[h]
    \caption{Inception Scores (IS), Fr\'{e}chet Inception Distances (FID) and the maximum intra FID (max-FID). Our proposed adaptive P2GAN achieves the highest IS, lowest FID and lowest max-FID in most cases. The top-two best performing methods are marked in boldface.}
    \label{tab:compare}
    \centering
    \resizebox{1\textwidth}{!}{
    \begin{tabular}{lccccccccccccccc}
    \toprule
    \multicolumn{1}{l}{} & \multicolumn{3}{c}{{Proj-GAN}} & \multicolumn{3}{c}{{TAC-GAN*}} & \multicolumn{3}{c}{{$f\mhyphen\text{cGAN}$} (ours)} & \multicolumn{3}{c}{{P2GAN} (ours)} & \multicolumn{3}{c}{{P2GAN-w} (ours)} \\
    \cmidrule(lr){2-4} \cmidrule(lr){5-7} \cmidrule(lr){8-10} \cmidrule(lr){11-13} \cmidrule(lr){14-16}
     & IS $\uparrow$ &  FID $\downarrow$ & max-FID $\downarrow$  & IS $\uparrow$ &  FID $\downarrow$ & max-FID $\downarrow$ & IS $\uparrow$ &  FID $\downarrow$ & max-FID $\downarrow$ & IS $\uparrow$ &  FID $\downarrow$ & max-FID $\downarrow$ & IS $\uparrow$ &  FID $\downarrow$ & max-FID $\downarrow$ \\
    \hline
    CIFAR100 &  {\bf 9.17 $\pm$ 0.14} &  9.98  & {\bf 186.46}   &  {\bf 8.86 $\pm$ 0.12} & {\bf 9.63} & 223.24  &  8.41 $\pm$ 0.09  & 10.76 & 215.86  &  8.55 $\pm$ 0.12 & 10.56 & 208.47  &  8.50 $\pm$ 0.10 & {\bf 9.84} & {\bf 204.69} \\
    ImageNet & 16.14 $\pm$ 0.34 & 22.26 & 147.69  &  14.85 $\pm$ 0.27 & 22.62 & 257.45  &  16.44 $\pm$ 0.26 & {\bf 19.28} & {\bf 145.44}  &  {\bf 17.78 $\pm$ 0.41}  & 19.80 & 170.40  &  {\bf 17.40 $\pm$ 0.33} & {\bf 18.87}  & {\bf 136.91} \\
    VGGFace200 & 50.93 $\pm$ 0.86 & 61.43 & 239.61  &  40.78 $\pm$ 0.57 & 96.06 & 478.10  &  109.94 $\pm$ 1.15 & 29.54 & 215.50  &  {\bf 148.48 $\pm$ 2.87} & {\bf 20.70} & {\bf 209.86} & {\bf 171.31 $\pm$ 3.44} & {\bf 15.70} & {\bf 127.43} \\
    VGGFace500 & 126.19 $\pm$ 1.95 & 23.57 & 162.27  &  150.05 $\pm$ 2.31 & 19.30 & 233.00  &  175.59 $\pm$ 2.46 & 16.74 & {\bf 136.21}  &  {\bf 210.75 $\pm$ 1.87} & {\bf 12.09} & {\bf 130.58}  &  {\bf 182.91 $\pm$ 1.24} & {\bf 12.73} & 151.66 \\
    \hline
    Average Rank & 3.5 & 4 & 3 & 4 & 3.75 & 5 & 3.5 & 3.25 & 2.75 & {\bf 1.75} &  2.5 & 2.5 & 2.25 & {\bf 1.5} & {\bf 1.75} \\
    \bottomrule
    \end{tabular}}
\end{table*}

\begin{table}[h]
    \caption{IS and FID scores on ImageNet at 128 resulution and CIFAR10. We mark `*' to FID and IS values reported in StudioGAN~\cite{kang2020contragan}.}
    \vspace{-0.1in}
    \label{tab:i128_c10}
    \centering
    \scalebox{1}{
    \begin{tabular}{lcccc}
    \toprule
    \multicolumn{1}{l}{} & \multicolumn{2}{c}{{ImageNet $128 \times 128$}} & \multicolumn{2}{c}{{CIFAR10}} \\
    \cmidrule(lr){2-3} \cmidrule(lr){4-5}
     &  IS $\uparrow$ & FID $\downarrow$ & IS $\uparrow$ & FID $\downarrow$ \\
    \hline
    Proj-GAN  & 30.73 & 23.07 & *9.85 & *8.03 \\
    P2GAN     & {\bf 59.24} & {\bf 16.86} & 9.76 & 8.00 \\
    P2GAN-w   & 42.69 & 19.19 & {\bf 9.87} & {\bf 7.99} \\
    \bottomrule
    \end{tabular}}
\end{table}
\begin{table}[t]
    \caption{Min-LPIPS on VGGFace2 dataset.}
    \vspace{-0.1in}
    \label{tab:lpips}
    \centering
    \scalebox{1}{
    \begin{tabular}{lcc}
    \toprule
     & {VGGFace200} & {VGGFace500} \\
    \hline
    Proj-GAN           & 0.198 & 0.373 \\
    TAC-GAN*           & 0.009 & 0.162 \\
    $f$-{cGAN}         & 0.233 & 0.385 \\
    P2GAN              & 0.172 & 0.376 \\
    P2GAN-w            & {\bf 0.387} & {\bf 0.416} \\
    \bottomrule
    \end{tabular}}
    \vspace{-0.05in}
\end{table}
\begin{figure*}[h]
    \centering
    \includegraphics[width=0.98\linewidth]{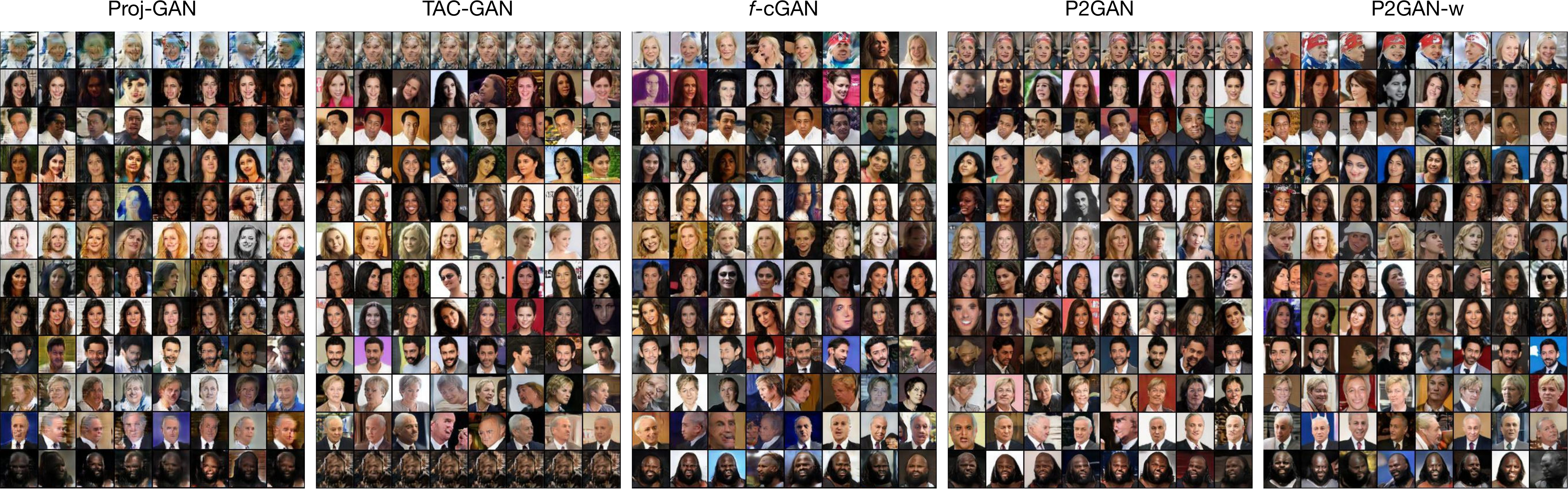}
    \caption{Samples of VGGFace200. Proj-GAN produces blurry faces. TAC-GAN and P2GAN occasionally show mode collapse (first and last row). The adaptive P2GAN model generates diverse and sharp faces.}
    \label{fig:sample_v200}
\end{figure*}
\begin{figure*}[h!]
    \centering
    \includegraphics[width=0.98\linewidth]{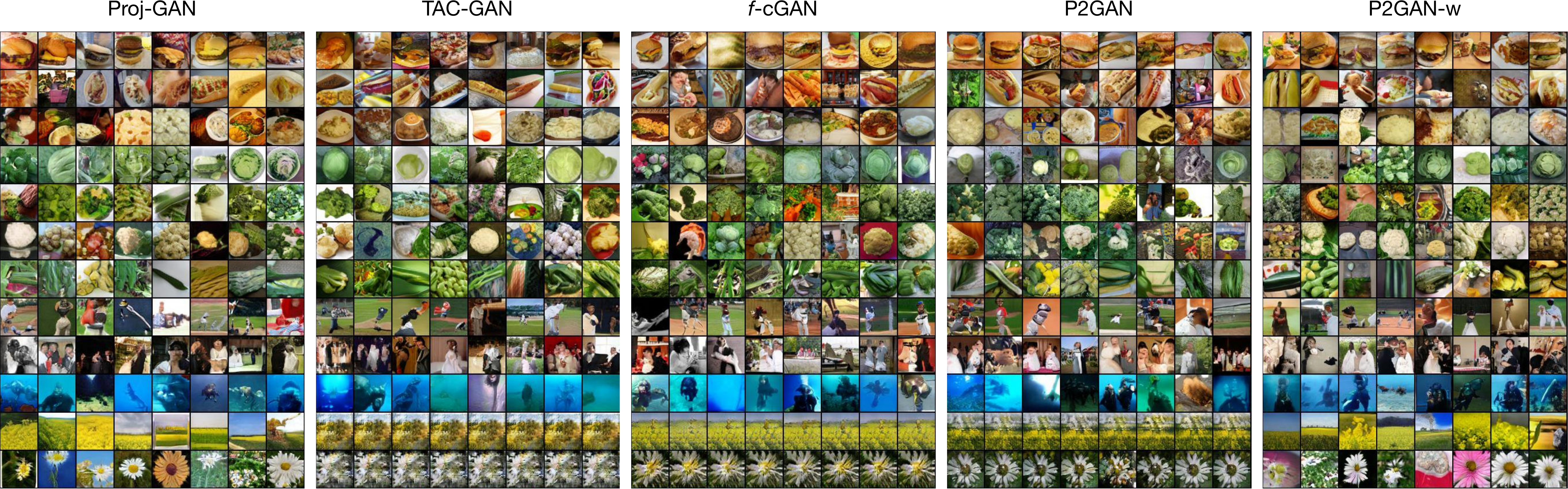}
    \caption{Samples of different methods trained on ImageNet. TAC-GAN, $f\mhyphen\text{cGAN}$ and P2GAN all show mode collapse (last two rows), while Proj-GAN and P2GAN-w generate diverse samples.}
    \label{fig:sample_i64}
\end{figure*}

\begin{figure*}[h]
  \begin{center}
    \hfill
    \subfloat[$\lambda$ on VGGFace200]{\includegraphics[width=0.22\linewidth]{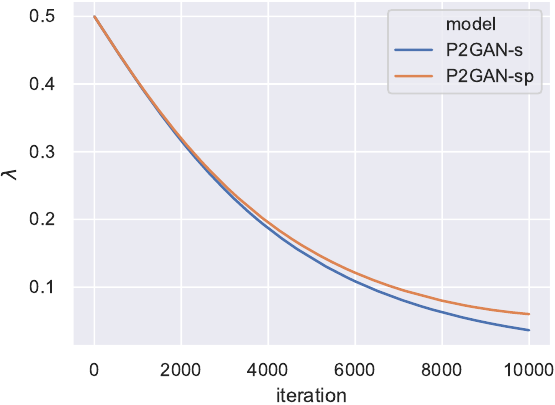}}\hfill
    \subfloat[$\lambda(x)$ on VGGFace200]{\includegraphics[width=0.22\linewidth]{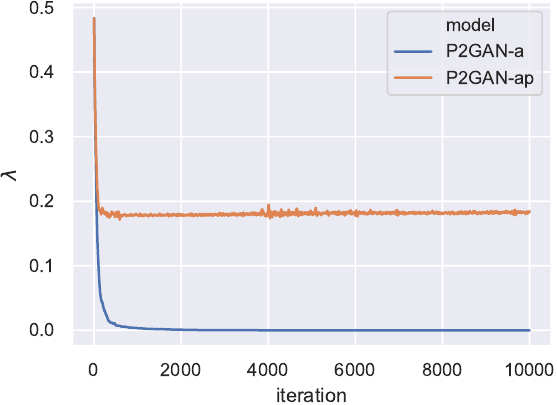}}\hfill
    \subfloat[$\lambda$ on CIFAR100IB]{\includegraphics[width=0.22\linewidth]{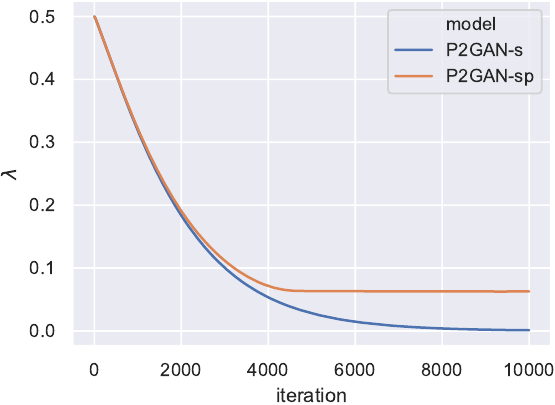}}\hfill
    \subfloat[$\lambda(x)$ on CIFAR100IB]{\includegraphics[width=0.22\linewidth]{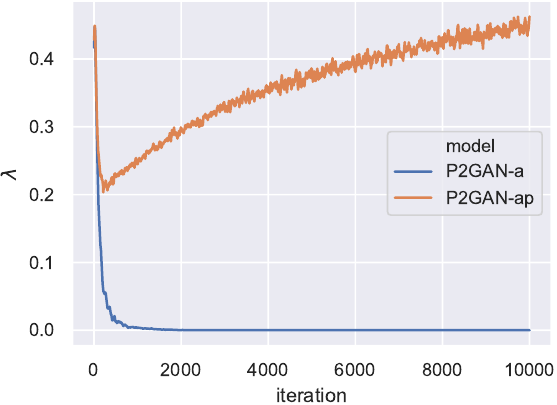}}\hfill
    \hfill
    \caption{Comparison of scalar $\lambda$ and amortized $\lambda(x)$. Plots show the value of $\lambda$ as training proceed. Curves are smoothed for better visualization.}
    \label{fig:weight}
  \end{center}
\end{figure*}

\begin{figure*}[h]
    \centering
    \includegraphics[width=0.9\linewidth]{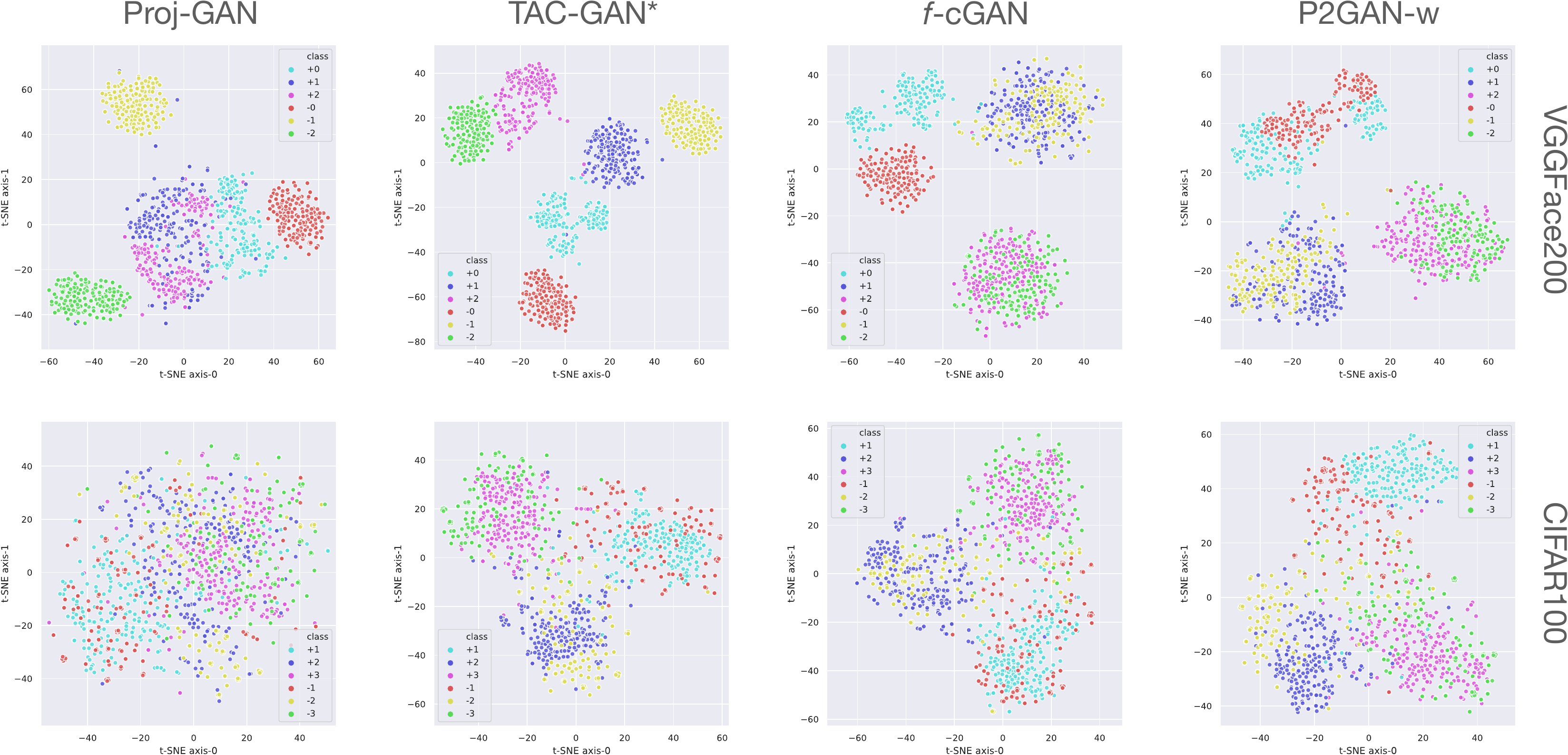}
    \caption{Data embedding visualized by t-SNE. Plot shows 2-D image embedding visualization for baselines at iteration 50000. Positive labels are real image embeddings and negative labels are fake image embeddings. A clear separation between real/fake indicates mode collapse.}
    \label{fig:tsne}
\end{figure*}

\section{Experiments}
\noindent \textbf{Datasets.}
In this section, we first show experimental results for model analysis on an imbalanced subset of CIFAR100~\cite{krizhevsky2009learning} (denoted as CIFAR100IB). Then we then report results of different baselines across various datasets. Finally, ablation studies are presented. We evaluate the distribution matching ability of different models on a synthetic Mixture of Gaussian (MoG) dataset~\cite{gong2019twin} and evaluate the image generation performance on CIFAR100, ImageNet~\cite{russakovsky2015imagenet} and VGGFace2~\cite{cao2018vggface2}. Following the protocol in~\cite{gong2019twin}, we construct 2 subsets VGGFace200 and VGGFace500.

\noindent \textbf{Baselines.} We compare our proposed models with competitive baselines including Proj-GAN and TAC-GAN. We use BigGAN~\cite{brock2018large} as backbone for baselines. For a fair comparison, we implemented TAC-GAN following the paper~\cite{gong2019twin}, and we denote as TAC-GAN* in following experiments. The code is written in PyTorch~\cite{paszke2019pytorch} and is available on the project website\footnote{\url{https://github.com/phymhan/P2GAN}}.

\noindent \textbf{Evaluation.}
Inception Scores (IS)~\cite{salimans2016improved} and Fr\'{e}chet Inception Distances (FID)~\cite{heusel2017gans} are reported for quantitative evaluation. We compute intra FIDs~\cite{miyato2018cgans} between the generated images and dataset images within each class. \cite{miyato2018cgans} report the mean value of intra FIDs while here we report the maximum among all classes to better capture mode collapse of trained generative models. Experimental setup and additional results are detailed in Supplementary.

\subsection{Model Analysis}
\noindent \textbf{Embedding visualization.}
We visualize the learned image embeddings of different methods, shown in Figure~\ref{fig:tsne}. For VGGFace200 dataset, we can see all baselines except for P2GAN-w show clear mode collapse (real and fake image embeddings are well separated) at iteration 50000. We can also observe that Proj-GAN does not enforce class separability. Moreover, comparing these two datasets, VGGFace is more clustered (in classes) than CIFAR100. This verifies our intuition that when the classification task is relatively easy, enforcing label matching helps the discriminator to be trained more powerful and thus provide more informative guidance to the generator.

\noindent \textbf{Learning adaptive weights.}
As shown in Figure~\ref{fig:weight}, without penalty, $\lambda$ and $\lambda(x)$ quickly converge to $0$ during training. Figure~\ref{fig:weight}-(d) shows an interesting phenomenon (from experiments on CIFAR100IB): The model quickly puts high weight on the discrimination loss $L_{D}$ at the early stage and weighs it less as training proceeds.

\noindent \textbf{$f\mhyphen\text{divergence}$.}
Here we consider several commonly used loss functions for $f\mhyphen\text{cGAN}$ and give the training results in Supplementary. We observe that the reverse-KL is the most stable and achieves highest IS and lowest FID. Thus, if not specified, reverse-KL is used in following experiments.


\subsection{Comparison of Baselines}
\noindent \textbf{Mixture of Gaussian (MoG).} We follow protocols in~\cite{gong2019twin} to set up experiments. The real data consists of three 1-D Gaussian with standard deviations $\sigma_0=1$, $\sigma_1=2$, and $\sigma_2=3$, and their means are equally spaced with interval $d_m$ ranging from $1$ to $5$. Average ranks of Maximum Mean Discrepancy (MMD)~\cite{gretton2012kernel} are reported in Table~\ref{tab:mog_rank}, where P2GAN achieves highest overall average rank. Detailed MMD metrics are reported in Supplementary. FID scores are reported in Table~\ref{tab:mog_fid}. P2GAN and $f\mhyphen\text{cGAN}$ shows advantages at lower distance values (more overlapping between modes), which suggests that if label matching can be effectively learned (the estimated posteriors are accurate), the discriminator is more powerful.

\noindent \textbf{CIFAR.}
Results on CIFAR100~\cite{krizhevsky2009learning} are reported in Table~\ref{tab:compare}. Proj-GAN achieves the lowest FID scores, while $f\mhyphen\text{cGAN}$ achieves the highest IS. The adaptive P2GAN model shows competitive high IS and low FID. It is also an interesting confirmation that the P2GAN-w indeed learns a model in between. Results on CIFAR10 are reported in Table~\ref{tab:i128_c10}.

\noindent \textbf{ImageNet.}
We conduct experiments on ImageNet~\cite{russakovsky2015imagenet} and compare with all baselines at resolution of $64 \times 64$. From Table~\ref{tab:compare}, we see P2GAN models achieve the highest IS and lowest FID. Intra FIDs are computed on 50 classes (samples of these 50 categories are given in Supplementary) due to its high computation cost. Figure~\ref{fig:sample_i64} shows that for the ``flower'' class in the last row, Proj-GAN and the proposed P2GAN-w generate diverse samples, while TAC-GAN, $f\mhyphen\text{cGAN}$ and P2GAN all show mode collapse. This can be verified by low max-FID values of the P2GAN-w model. Results of 128 resolution experiments are reported in Table~\ref{tab:i128_c10}.

\noindent \textbf{VGGFace2.}
VGGFace2~\cite{cao2018vggface2} is a large-scale dataset for face recognition, and contains over 9000 identities with around 362 images for each identity. The VGGFace2 dataset is introduced in \cite{gong2019twin} for image generation task. We follow their experimental setup and construct subsets VGGFace200 and VGGFace500. The images are resized to $64 \times 64$. Table~\ref{tab:compare} shows that the proposed $f\mhyphen\text{cGAN}$ and P2GAN models significantly outperform Proj-GAN and TAC-GAN. To directly measure diversity, we report the min-LPIPS (across all categories) in Table~\ref{tab:lpips}. From Figure~\ref{fig:sample_v200} we can see that Proj-GAN, TAC-GAN and P2GAN all show mode collapse to a certain degree. While P2GAN-w generates sharp and diverse images. Samples from Proj-GAN are also blurry. 

\subsection{Ablation Studies}
A na\"ive hybrid model could be an ensemble of the Proj-GAN loss and the $f\mhyphen\text{cGAN}$ loss, which requires three parameters $v_y$, $v^p_y$ and $v^q_y$. The resulting does not truly untie class embeddings and less elegant. We find it does not show superior performance over dual projection. We provide in Table~\ref{tab:ablation} comparisons of the na\"ive baseline, an over-parameterization baseline ($\lambda \equiv 0$), and DM-GAN ($\psi \equiv 0$). Also, the training curves on VGGFace200 given in Supplementary shows that over-parameterization alone does not prevent Proj-GAN from failing on this dataset and dual projections indeed benefit from balanced data matching and label matching.
\begin{table}[h!]
    \caption{FID scores of ablation studies. All models are trained for 60000 iterations on ImageNet, 62000 iterations on CIFAR100, and 50000 iterations on VGGFace200.}
    \label{tab:ablation}
    \centering
    \scalebox{1}{
    \begin{tabular}{lccc}
    \toprule
                & ImageNet & CIFAR100 & VGGFace200 \\
    \hline
    Proj-GAN            & 24.57 & 10.02 & 61.27 \\
    P2GAN               & 21.46 & 10.55 & 23.15 \\
    $\lambda \equiv 0$  & 23.64 & 10.82 & 62.46 \\
    $\psi \equiv 0$     & 26.30 & 10.95 & 87.01 \\
    Na\"ive             &  -    & 10.58 & 32.31 \\
    \bottomrule
    \end{tabular}}
\end{table}

\section{Conclusion}
In this paper, we give insights on projection form of conditional discriminators and propose a new conditional generative adversarial network named Dual Projection GAN (P2GAN). We demonstrate its flexibility in modeling and balancing data matching and label matching. We further rigorously analyze the underlying connections between AC-GAN, TAC-GAN, and Proj-GAN. From the analysis, we first propose $f\mhyphen\text{cGAN}$, a general framework for learning conditional GAN. We demonstrate the efficacy of our proposed models on various synthetic and real-world datasets. P2GAN may also be potentially applicable to image-to-image translation~\cite{isola2017image,park2019semantic} and federated learning~\cite{liu2021fedlearn} which we leave for future work.

\subsection*{Acknowledgments}
This research was funded based on partial funding from NSF: IIS 1703883, NSF IUCRC CNS-1747778, CCF-1733843, IIS-1763523, IIS-1849238, and NIH: 1R01HL127661-01 and NIH: R01HL127661-05.

{\small
\bibliographystyle{ieee_fullname}
\bibliography{egbib}
}

\newpage
\section*{Supplementary}
\section{Proof of Proposition \ref{prop1}}
\begin{proposition}
When $\psi=0$, a Proj-GAN reduces to $K$ unconditional GANs, each of them minimizes the Jensen-Shannon divergence between $P_{X|y}$ and $Q_{X|y}$ with mixing ratio $\{\frac{P(y)}{P(y)+Q(y)}, \frac{Q(y)}{P(y)+Q(y)}\}$. Its value function can be written as,
\begin{align}
    \mathbb{E}_{P_Y}\left\{ \mathbb{E}_{P_{X|Y}}\log{D(x|y)}+\frac{Q_y}{P_y}\mathbb{E}_{Q_{X|Y}}\log{(1-D(x|y))} \right\}.\nonumber
\end{align}
\label{prop1_supp}
\end{proposition}
\begin{proof}
When $\psi(\cdot)$ is zero, $\tilde{D}(x,y)=v_y^\text{T}\phi(x)$. Recall the logit of an unconditional GAN is $\tilde{D}(x)=v_X^\text{T}\phi(x)$ (with bias $b=0$). It immediately follows that matrix $V$ is a collection of $K$ vectors $v_y$, one for each class. Simply rearranging the cGAN objective, we get
\begin{align}
    \mathbb{E}_{P(y)}\left\{ \mathbb{E}_{P(x|y)}\log{D(x|y)}+\frac{1}{r(y)}\mathbb{E}_{Q(x|y)}\log{(1-D(x|y))} \right\},
\end{align}
\noindent with $r(y)=\frac{P(y)}{Q(y)}$. This can be viewed as a weighted sum of $K$ GAN objectives with binary cross-entropy loss. Each of them minimizes the Jensen-Shannon divergence between $P_{X|y}$ and $Q_{X|y}$ with weights $\{\frac{P(y)}{P(y)+Q(y)}, \frac{Q(y)}{P(y)+Q(y)}\}$.
\end{proof}

\section{Proof of Proposition \ref{prop2}}
\setcounter{theorem}{0}
\begin{lemma}
For any classifier $C$, the objective $\mathbb{E}_{x,y \sim P_{XY}}{\log C(x,y)} \leq -H_P(Y|X)$, and the maximizer $C^*$ is obtained if and only if $Q^{c}(y|x)=P(y|x)$, where $Q^c$ is the conditional distribution induced by $C$.
\label{lemma1}
\end{lemma}
\begin{proof}
It follows immediately with the observation,
\begin{align}
    {L}_\text{CE} =& \mathbb{E}_{x,y \sim P_{XY}}{\log {C}(x,y)} \nonumber \\
    =& \mathbb{E}_{x \sim P_X}{\mathbb{E}_{y \sim P_{Y|X}}{\log P(y|x)\frac{Q^{c}(y|x)}{P(y|x)}}} \nonumber \\
    =& \mathbb{E}_{x \sim P_X}{\mathbb{E}_{y \sim P_{Y|X}}{\log P(y|x)}} - \nonumber \\
     & \mathbb{E}_{x \sim P_X}{\text{KL}(P_{Y|X} \Vert Q^{c}_{Y|X})} \nonumber \\
    \leq & -H_P(Y|X).
\end{align}
The equality is achieved if and only if $Q^{c}_{Y|X}=P_{Y|X}$.
\end{proof}
\begin{proposition}
Given a generator $G$, if cross entropy losses $L^{p}_{mi}$ and $L^{q}_{mi}$ are minimized optimally, then the difference of two losses evaluated at fake data equals the reverse KL-divergence between $P_{Y|X}$ and $Q_{Y|X}$,
\begin{align}
    L^{p}_{mi}(x^-)-L^{q}_{mi}(x^-) = \mathbb{E}_{Q_{X}}{\text{KL}(Q_{Y|X} \Vert P_{Y|X})}.
\end{align}
\label{prop2}
\end{proposition}
\begin{proof}
Applying Lemma \ref{lemma1} to classifier $C^p$ and $C^q$ respectively, we have
\begin{align}
    C^{p*}=P(y|x) \quad \text{and} \quad C^{q*}=Q(y|x).
\end{align}
\noindent Then,
\begin{align}
    &L^{p*}_{mi}(x^-)-L^{q*}_{mi}(x^-) \nonumber \\
    = &\mathbb{E}_{z \sim P_Z, y \sim P_Y}{ \log Q^{cq*}(G(z,y),y) - \log Q^{cp*}(G(z,y),y) } \nonumber \\
    = &\mathbb{E}_{z \sim P_Z, y \sim P_Y}{ \log \frac{C^{q*}(G(z,y),y)}{C^{p*}(G(z,y),y)} } \nonumber \\
    = &\mathbb{E}_{Q_{X}}{ \log \frac{Q(y \vert x)}{P(y \vert x)} } \nonumber \\
    = &\mathbb{E}_{Q_{X}}{\text{KL}(Q_{Y|X} \Vert P_{Y|X})}.\nonumber
\end{align}
\end{proof}

\section{Proof of Theorem \ref{thm1}}
\setcounter{theorem}{0}
\begin{theorem}
Denoting $P_{XY}$ and $Q_{XY}$ as the data distribution and the distribution induced by $G$, their Jensen-Shannon divergence is upper bounded by the following,
\begin{align}
    & \text{JSD}(P_{XY},Q_{XY}) \le  \\
    & 2c_1 \sqrt{2 \text{JSD}(P_X, Q_X)} + c_2 \sqrt{2 \text{KL}(P_{Y|X} \Vert Q^{p}_{Y|X})} + \nonumber\\
    & c_2 \sqrt{2 \text{KL}(Q_{Y|X} \Vert Q^{q}_{Y|X})} + c_2 \sqrt{2 \text{KL}(Q^{q}_{Y|X} \Vert Q^{p}_{Y|X})}.\nonumber
\end{align}
\label{thm1_supp}
\end{theorem}

\begin{proof}
According to the triangle inequality of the total variation distance (TV, denoted as $\delta$), we have
\begin{align}
    & \delta(P_{XY}, Q_{XY}) \nonumber \\
    \le & \underbrace{ \delta(P_{XY}, P_{Y|X} Q_{X}) }_{\textcircled{\scriptsize{I}}} + \underbrace{ \delta(P_{Y|X} Q_{X}, Q_{XY}) }_{\textcircled{\scriptsize{II}}}. \label{eq:thm1_1}
\end{align}
\noindent We can relax term \textcircled{\scriptsize{I}} using the definition of TV,
\begin{align}
     & \delta(P_{XY}, P_{Y|X} Q_{X}) = \delta(P_{Y|X} P_{X}, P_{Y|X} Q_{X}) \nonumber \\
    =& \frac{1}{2} \int{ \Large{\{} |P_{Y|X}(y|x)P_{X}(x) -  P_{Y|X}(y|x)Q_{X}(x)|\mu(x,y) \Large{\}} } \nonumber \\
    \overset{\mathrm{(a)}}{\le} & \frac{1}{2} \int |P_{Y|X}(y|x)| \mu(x,y) \int |P_X(x)-Q_X(x)| \mu(x,y) \nonumber \\
    \le & c_1 \delta(P_{X}, Q_{X}), \label{eq:thm1_2}
\end{align}
\noindent where $\mu$ is a ($\sigma$-finite) measure, $c_1$ is an upper bound of $\int |P_{Y|X}(y|x)| \mu(x,y)$. (a) follows from the H\"older inequality.
Similarly, for \textcircled{\scriptsize{II}} we have,
\begin{align}
    \delta(P_{Y|X} Q_{X}, Q_{XY}) =& \delta(P_{Y|X} Q_{X}, Q_{Y|X} Q_{X}) \nonumber \\
    \le & c_2 \delta(P_{Y|X}, Q_{Y|X}), \label{eq:thm1_3}
\end{align}
\noindent and $c_2$ is an upper bound of $\int |Q_{X}(x)| \mu(x)$.

\noindent Then, using the triangle inequality of TV again,
\begin{align}
    & \delta(P_{Y|X}, Q_{Y|X}) \label{eq:thm1_4} \\
   \le & \delta(P_{Y|X}, Q^{p}_{Y|X}) + \delta(Q^{p}_{Y|X}, Q^{q}_{Y|X}) + \delta(Q^{q}_{Y|X}, Q_{Y|X}). \nonumber
\end{align}
\noindent Combining Equation~\ref{eq:thm1_1}, \ref{eq:thm1_2}, \ref{eq:thm1_3} and \ref{eq:thm1_4},
\begin{align}
    & \delta(P_{XY}, Q_{XY}) \nonumber \\
    \le & c_1 \delta(P_{X}, Q_{X}) + c_2 \delta(P_{Y|X}, Q_{Y|X}) \nonumber \\
    \le & \underbrace{ c_1 \delta(P_{X}, Q_{X}) }_{\textcircled{\scriptsize{III}}} + \underbrace{ c_2 \delta(Q^{p}_{Y|X}, Q^{q}_{Y|X}) }_{\textcircled{\scriptsize{IV}}} + \nonumber \\
    & \underbrace{ c_2 \delta(P_{Y|X}, Q^{p}_{Y|X}) + c_2 \delta(Q^{q}_{Y|X}, Q_{Y|X}) }_{\textcircled{\scriptsize{V}}}.
\end{align}
\noindent From above, we see that \textcircled{\scriptsize{III}} is enforced by the unconditional GAN, \textcircled{\scriptsize{IV}} is minimized by the $f$-divergence term, and \textcircled{\scriptsize{V}} is bounded by $L^p_{mi}$ and $L^q_{mi}$.

\noindent Finally, using to Pinsker inequality~\cite{tsybakov2008introduction} $\delta(P, Q) \le \sqrt{\frac{1}{2}KL(P \Vert Q)}$, and Lemma 3 in~\cite{thekumparampil2018robustness} $\frac{1}{2}\delta^2(P, Q)\le JSD(P,Q) \le 2 \delta(P, Q)$, we have,
\begin{align}
    & \quad \text{JSD}(P_{XY},Q_{XY}) \le  \\
    & 2c_1 \sqrt{2 \text{JSD}(P_X,Q_X)} + c_2 \sqrt{2 \text{KL}(Q^{q}_{Y|X} \Vert Q^{p}_{Y|X})} + \nonumber\\
    & c_2 \sqrt{2 \text{KL}(P_{Y|X} \Vert Q^{p}_{Y|X})} + c_2 \sqrt{2 \text{KL}(Q_{Y|X} \Vert Q^{q}_{Y|X})}.\nonumber
\end{align}
\end{proof}

\section{Weighted Dual Projection GAN}
\noindent \textbf{P2GAN-ap.} The full objectives of P2GAN with amortised weights are as follows,
\begin{align}
    L_{D}^{P2ap} = &\mathbb{E}_{x,y \sim P_{XY}}{(1-\lambda(x))\mathcal{A}(-\tilde{D}(x,y))}+ \label{eq:p2ap_full} \\
    & \mathbb{E}_{z \sim P_Z, y \sim Q_Y}{(1-\lambda(G(z,y)))\mathcal{A}(\tilde{D}(G(z,y),y))}- \nonumber \\
    & \mathbb{E}_{x,y \sim P_{XY}}{\lambda(x)T^p(x,y)}- \nonumber \\
    & \mathbb{E}_{z \sim P_Z, y \sim Q_Y}{\lambda(G(z,y))T^q(G(z,y),y)} \quad \text{and} \nonumber \\
    L_{G}^{P2ap} = &\mathbb{E}_{z \sim P_Z, y \sim Q_Y}{(1-\lambda(G(z,y)))\mathcal{A}(-\tilde{D}(G(z,y),y))}. \nonumber
\end{align}
\noindent Here $T^p$ and $T^q$ has the same definition as in $f\mhyphen\text{cGAN}$.

\noindent \textbf{Alternative weighing strategies.} An alternative design of a weighted P2GAN is to fix the weight of $L_D$ to $1$,
\begin{align}
    L^{P2sp \mhyphen alt}_{D} = L_{D} + \lambda \cdot (L^p_{mi}+L^q_{mi}) - \frac{1}{2}\log{\lambda}.\label{eq:p2sp_alt}
\end{align}
\noindent Here, $\lambda \in [0, \infty)$ and is initialized as $1$. We can define similar alternatives for P2GAN-s, P2GAN-a and P2GAN-ap. The key difference is that, weighing $(1-\lambda) \cdot L_D$ and $\lambda \cdot L_{mi}$ explicitly balances data matching and label matching, while the alternative way balances $L_{mi}$ and the penalty term. Without penalty, $\lambda$ in all alternative variants will vanish since this minimizes the total loss, however, the decreasing rate is determined by loss $L_{mi}$ adaptively. An extended comparison is listed in Table~\ref{tab:p2w_alt}. In practice, we find the differences are not significant.

\begin{table}[h!]
    \caption{FID scores of alternative weighting strategies for P2GAN-w. All models are trained for 62000 iterations on CIFAR100 and 50000 iterations on VGGFace200.}
    \label{tab:p2w_alt}
    \centering
    \begin{tabular}{lcc}
    \toprule
             & CIFAR100 & VGGFace200 \\
    \hline
    P2GAN-s          & {\bf 9.03} & 20.59 \\
    P2GAN-sp         & 9.51 & 20.18 \\
    P2GAN-a          & 10.13 & 20.26 \\
    P2GAN-ap         & 9.82 & {\bf 18.99} \\
    \hline
    P2GAN-s-alt      & {\bf 9.49} & {\bf 20.21} \\
    P2GAN-sp-alt     & 9.85 & 20.82 \\
    P2GAN-a-alt      & 9.98 & 23.25 \\
    P2GAN-ap-alt     & 9.72 & 21.57 \\
    \bottomrule
    \end{tabular}
\end{table}

\section{$f\mhyphen\text{divergence}$}
\begin{figure}[h]
  \begin{center}
    \subfloat[IS on CIFAR100IB]{\includegraphics[width=0.48\linewidth]{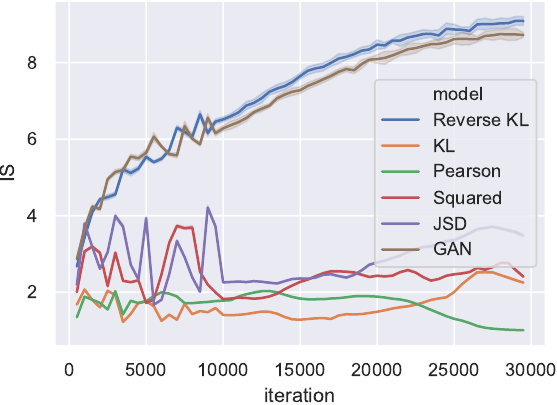}}\hfill
    \subfloat[FID on CIFAR100IB]{\includegraphics[width=0.48\linewidth]{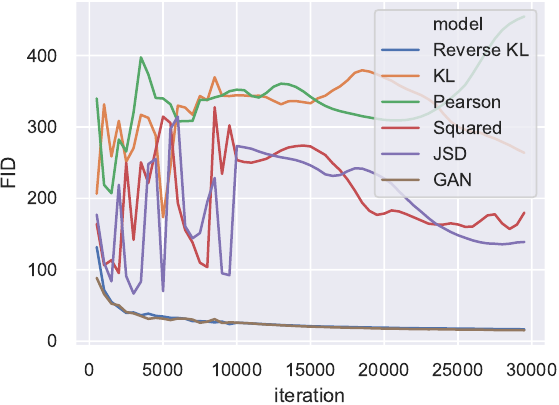}}\par
    \subfloat[IS on VGGFace200]{\includegraphics[width=0.48\linewidth]{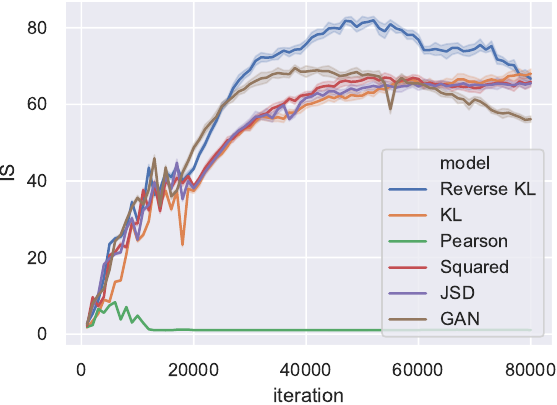}}\hfill
    \subfloat[FID on VGGFace200]{\includegraphics[width=0.48\linewidth]{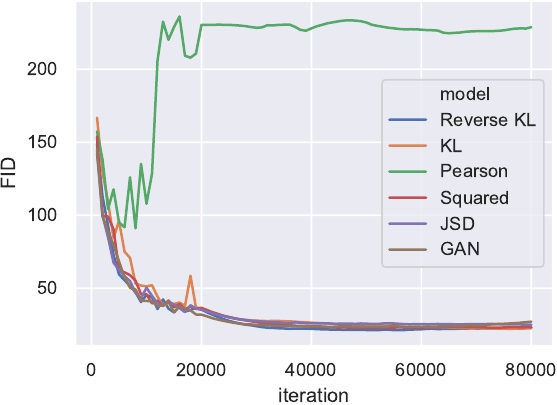}}
    \caption{(a-b) IS and FID on CIFAR100IB. (c-d) IS and FID on VGGFace200. Different curves correspond to different choices of $f\mhyphen\text{divergence}$, CE loss is used for both $P$ and $Q$.}
    \label{fig:fdiv}
  \end{center}
\end{figure}

Here we consider several $f\mhyphen\text{divergence}$ loss functions~\cite{nowozin2016f} and list them in Table~\ref{tab:f_loss}. Results on CIFAR100IB and VGGFace200 is given in Figure~\ref{fig:fdiv}. Different from the results on VGGFace200, only reverse-KL and GAN losses are stable on CIFAR100IB.
\begin{table*}[h]
    \caption{List of $f\mhyphen\text{divergence}$ and their corresponding {\it generator function} $f(\cdot)$}
    \label{tab:f_loss}
    \centering
    \begin{tabular}{lcc}
    \toprule
    Name             & $f(u)$           & $f\circ\exp{(u)} $ \\
    \hline
    Reverse KL       & $-\log{u}$       & $-u$            \\
    Kullback-leibler & $u\log{u}$       & $u {e^u}$     \\
    Pearson $\chi^2$ & $(u-1)^2$        & $({e^u}-1)^2$ \\
    Squared Hellinger& $(\sqrt{u}-1)^2$ & $(e^{u/2}-1)^2$ \\
    Jensen-Shannon   & $-(u+1)\log\frac{1+u}{2}+u\log{u}$ & $-({e^u}+1)\log\frac{1+{e^u}}{2}+u {e^u}$\\
    GAN              & $u \log{u}-(u+1)\log{(u+1)}$ & $u e^{u}-(e^{u}+1)\log{(e^{u}+1)}$ \\
    \bottomrule
    \end{tabular}
\end{table*}
\begin{figure}[h!]
  \begin{center}
    \subfloat[IS on ImageNet]{\includegraphics[width=0.48\linewidth]{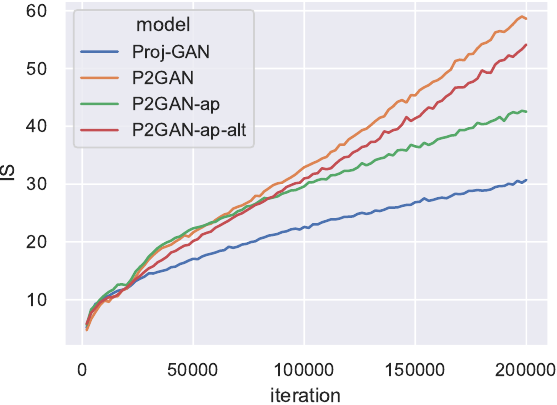}}
    \subfloat[FID on ImageNet]{\includegraphics[width=0.48\linewidth]{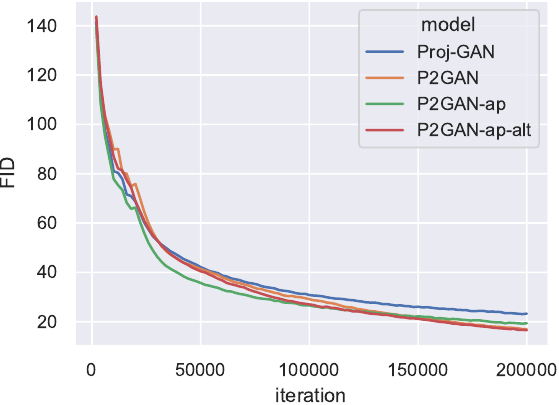}}
    \caption{IS and FID score over iterations on ImageNet at at $128 \times 128$ resolution.}
    \label{fig:train_i128}
  \end{center}
\end{figure}
\begin{figure}[h!]
  \begin{center}
    \subfloat[IS on VGGFace200]{\includegraphics[width=0.48\linewidth]{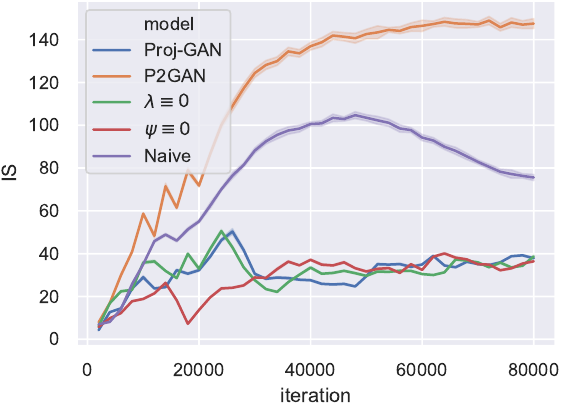}}
    \subfloat[FID on VGGFace200]{\includegraphics[width=0.48\linewidth]{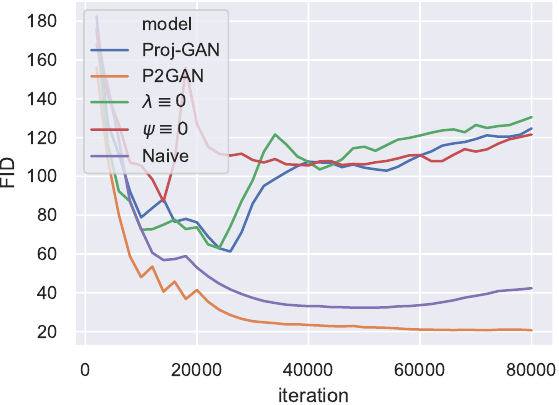}}
    \caption{IS and FID score over iterations on VGGFace200.}
    \label{fig:train_v200}
  \end{center}
\end{figure}
\begin{table}[h!]
    \caption{Inception Scores (IS), Fr\'{e}chet Inception Distances (FID) and the maximum intra FID (max-FID), evaluated on VGGFace200 dataset.}
    \label{tab:collapse_v200}
    \centering
    \begin{tabular}{lccc}
    \toprule
                     & IS $\uparrow$ & FID $\downarrow$ & max-FID $\downarrow$ \\
    \hline
    Proj-GAN               &  50.93 $\pm$ 0.86 & 61.43 & 239.61 \\
    TAC-GAN*               &  40.78 $\pm$ 0.57 & 96.06 & 478.10 \\
    Na\"ive                & 104.52 $\pm$ 1.95 & 32.39 & 196.99 \\
    $f\mhyphen\text{cGAN}$ & 109.94 $\pm$ 1.15 & 29.54 & 215.50 \\
    P2GAN                  & 148.48 $\pm$ 2.87 & 20.70 & 209.86 \\
    P2GAN-w                & {\bf 171.31 $\pm$ 3.44} & {\bf 15.70} & {\bf 127.43} \\
    \bottomrule
    \end{tabular}
\end{table}

\section{Implementation}
The 1D Mixture of Gaussian experiments are implemented based on the official TAC-GAN repo\footnote{\url{https://github.com/batmanlab/twin-auxiliary-classifiers-gan}}. Code for CIFAR100, VGGFace2, and ImageNet at resolution $64 \times 64$ are written based on the BigGAN-PyTorch repo\footnote{\url{https://github.com/ajbrock/BigGAN-PyTorch}}. Code for CIFAR10 and ImageNet at resolution $128 \times 128$ is implemented based on StudioGAN~\cite{kang2020contragan} repo\footnote{\url{https://github.com/POSTECH-CVLab/PyTorch-StudioGAN}}.

\section{1D MoG Synthetic Data}
\noindent \textbf{Experimental setup.} We follow the same protocol as in TAC-GAN paper~\cite{gong2019twin}. The standard deviations $\sigma_0=1$, $\sigma_1=2$, and $\sigma_2=3$ are fixed, and distance $d_m$ is set to value $1, 2, \ldots, 5$ and all models are trained 100 times for each experimental setting. The code for synthetic data, network architectures, and MMD evaluation metrics are borrowed from the official TAC-GAN repo. However, the training code for hinge loss is not provided, thus we implemented our hinge loss version based on the BigGAN-PyTorch repo.

\begin{figure*}[t]
    \centering
    \includegraphics[width=0.9\linewidth]{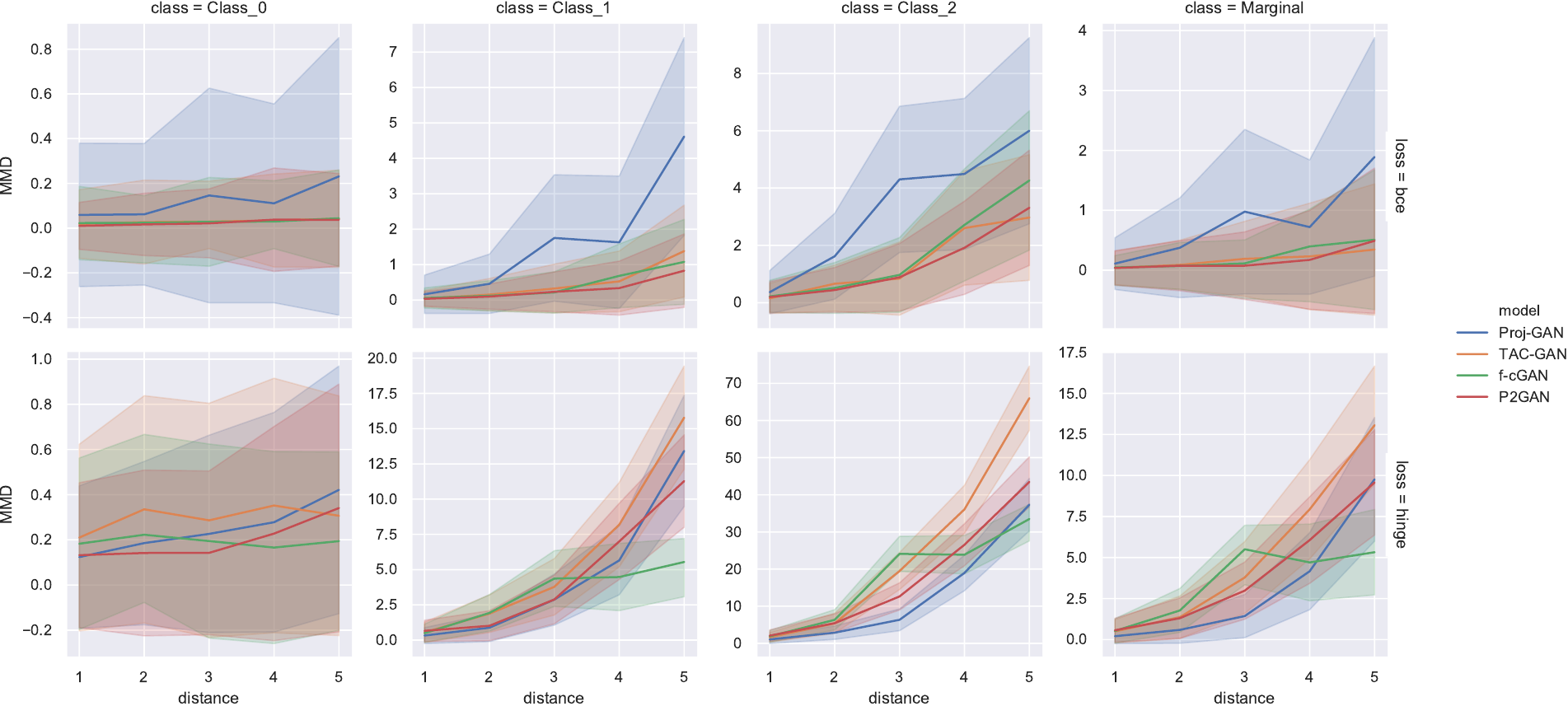}
    \caption{The Maximum Mean Discrepancy (MMD) metric. Proposed methods show low MMD with low variance across different runs.  }
    \label{fig:mog_mmd}
\end{figure*}

\begin{figure*}[t]
  \begin{center}
    \subfloat[distance $d_m=1$]{\includegraphics[width=0.48\linewidth]{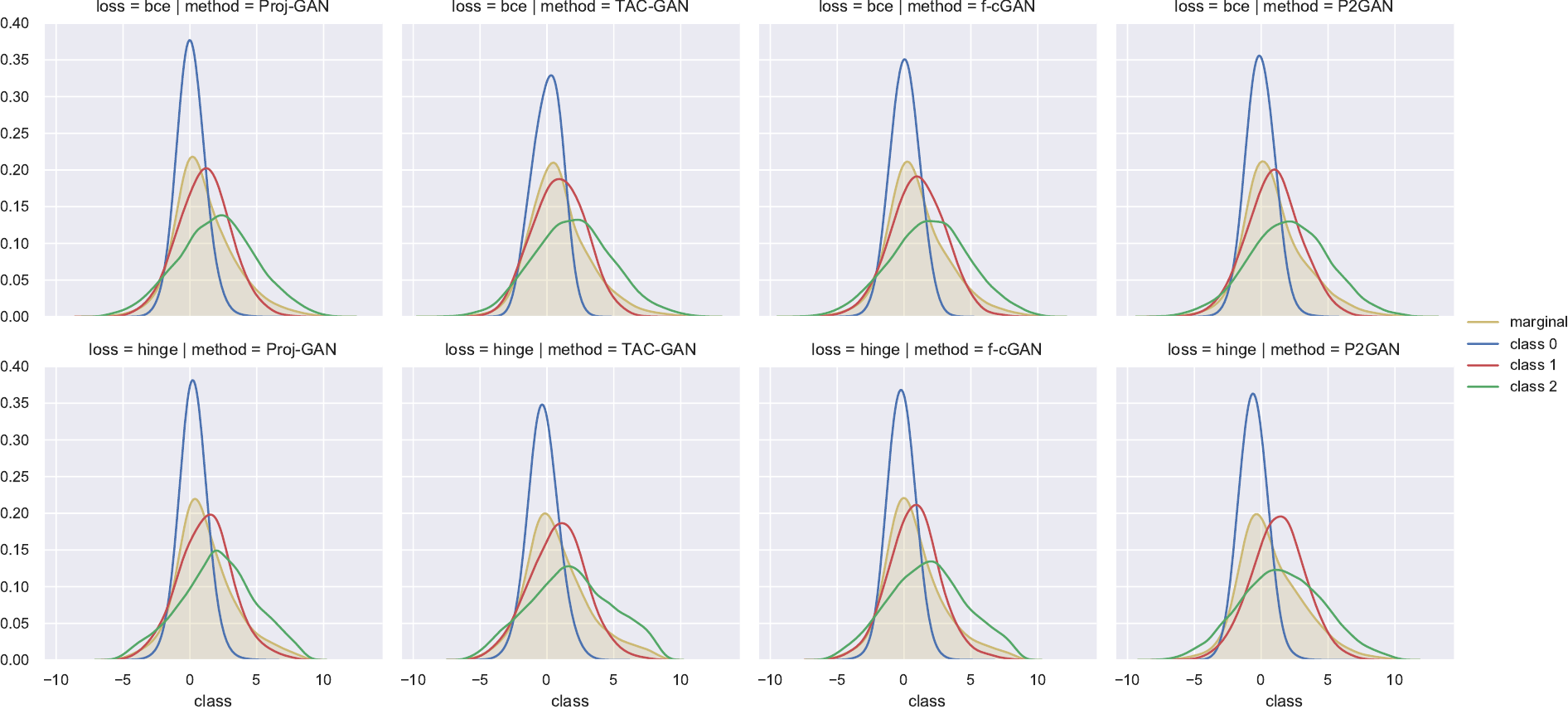}}\hfill
    \subfloat[distance $d_m=2$]{\includegraphics[width=0.48\linewidth]{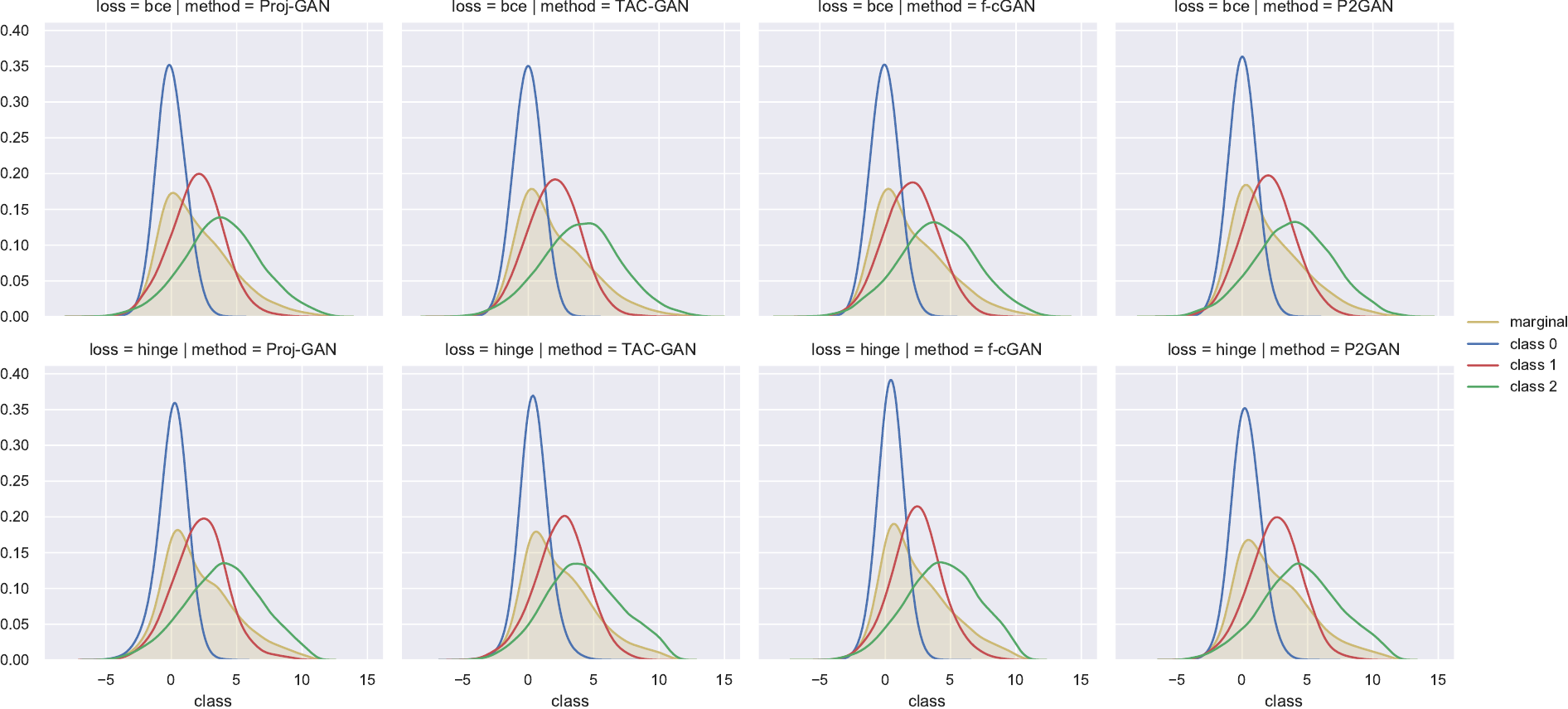}}\par
    \subfloat[distance $d_m=3$]{\includegraphics[width=0.48\linewidth]{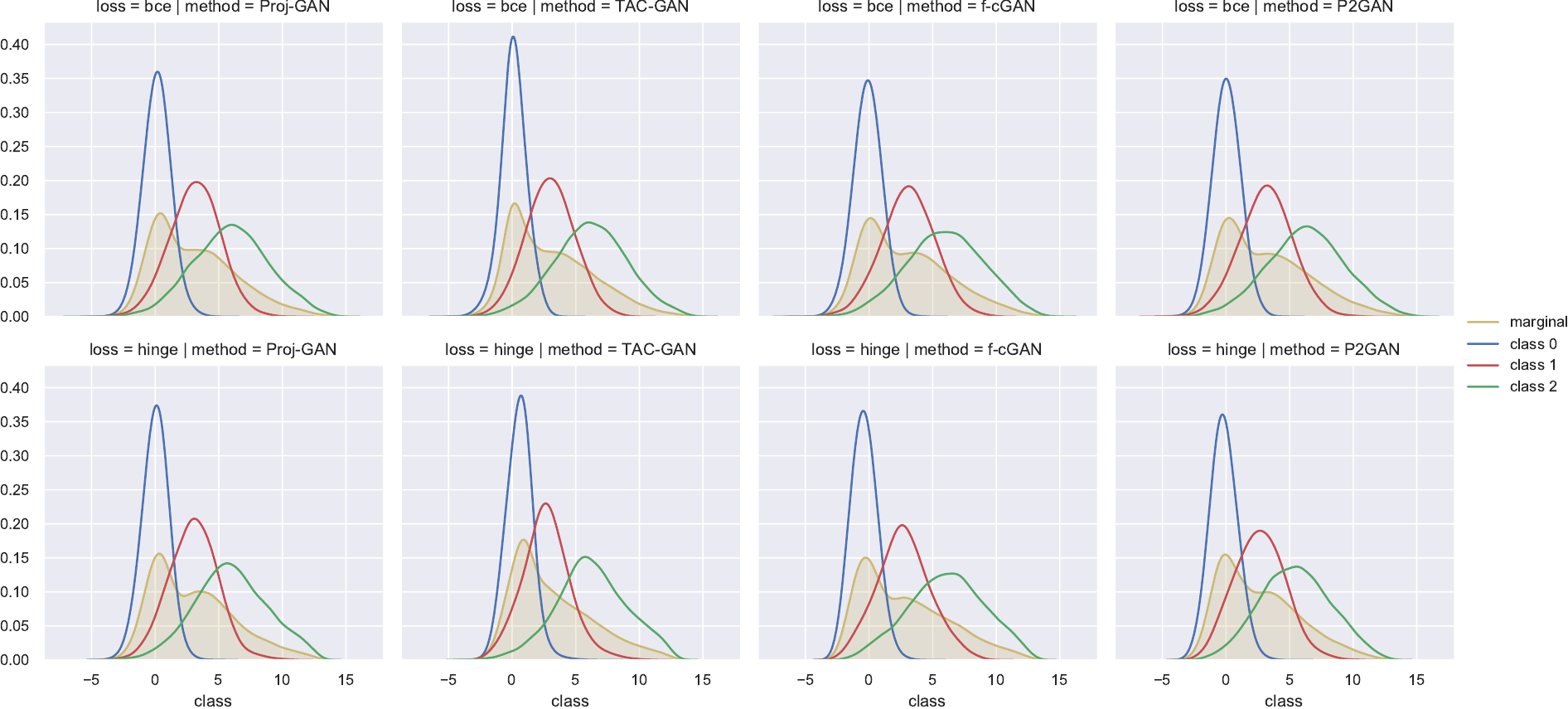}}\hfill
    \subfloat[distance $d_m=4$]{\includegraphics[width=0.48\linewidth]{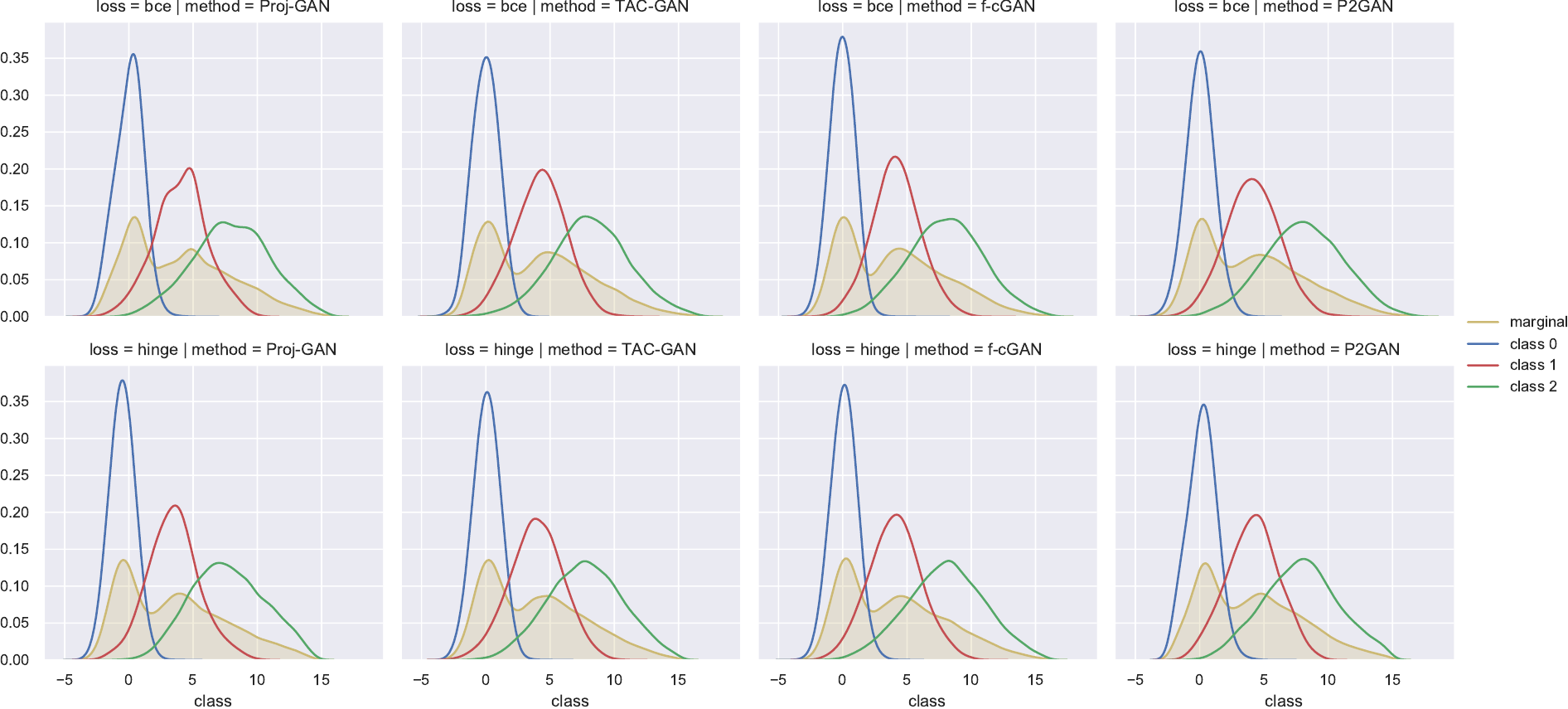}}\par
    \subfloat[distance $d_m=5$]{\includegraphics[width=0.48\linewidth]{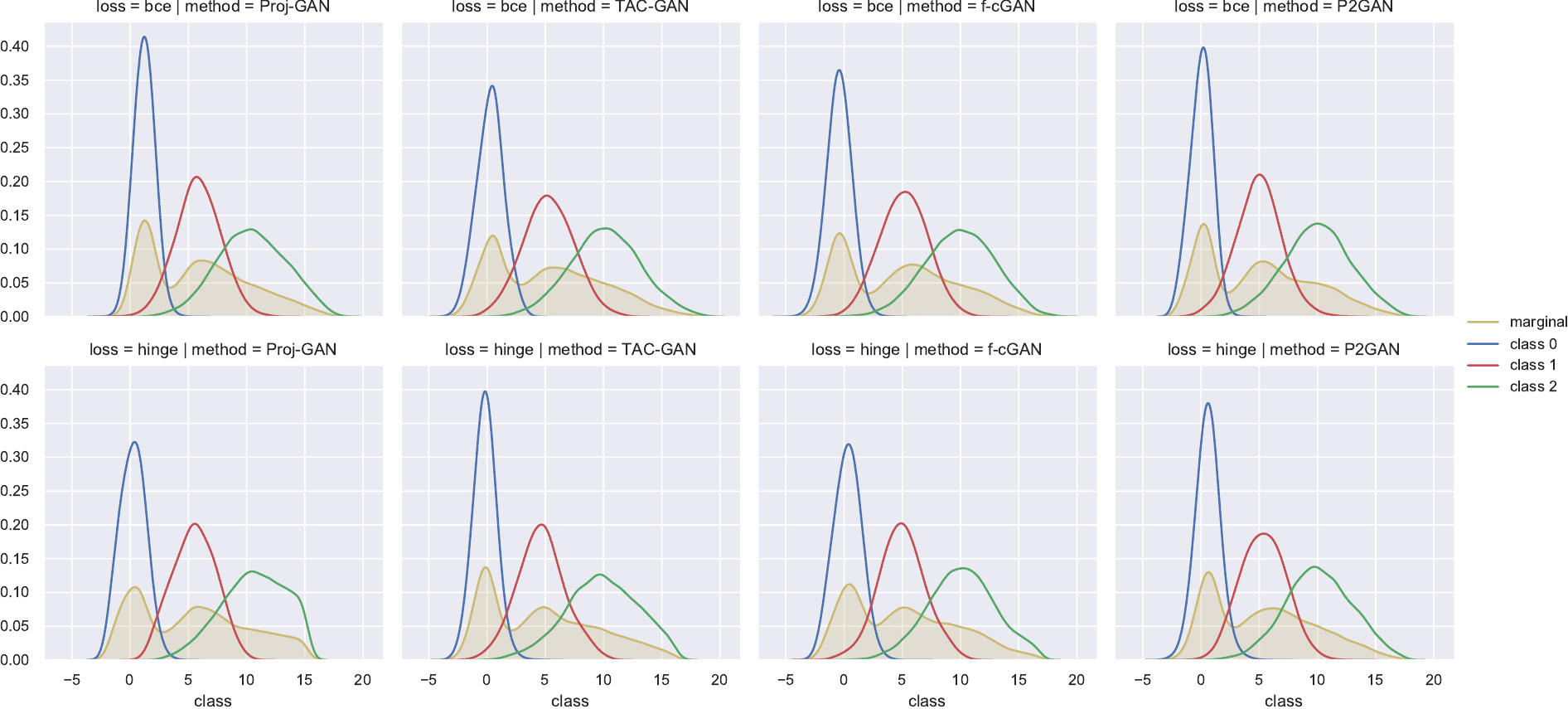}}
    \caption{Change distance $d_m$ between the means of adjacent 1-D Gaussian components. For each sub-figure, the first row adopts binary cross entropy loss and the second row adopts hinge loss.}
    \label{fig:mog_plot}
  \end{center}
\end{figure*}

\noindent \textbf{More results.} The average MMD values across 100 runs are reported in Table~\ref{tab:mog_mmd} and Figure~\ref{fig:mog_mmd}. Samples of generated 1D MoG are visualized in Figure~\ref{fig:mog_plot}. We observe that P2GAN performs the best with BCE loss, demonstrating its ability to generate accurate distributional data. Even with hinge loss, P2GAN still performs relatively well, and achieves the highest overall ranking.

\begin{table*}[t]
    \caption{The Maximum Mean Discrepancy (MMD) metric on 1D Mixture of Gaussian (MoG) synthetic dataset. Classes `0', `1', `2' stand for mode 0, 1, 2, and `M' stands for marginal. The upper half lists results of BCE loss and the lower half lists results when adopting hinge loss. We run each experiment 100 times and report the average MMD over the top $90\%$ performing runs. Standard deviations are omitted due to space limit. Entries with two lowest values are marked in boldface.}
    \label{tab:mog_mmd}
    \centering
    \resizebox{1\linewidth}{!}{
    \begin{tabular}{lcccccccccccccccccccc}
    \toprule
    {BCE~/~Hinge} & \multicolumn{4}{c}{$d_m=1$} & \multicolumn{4}{c}{$d_m=2$} & \multicolumn{4}{c}{$d_m=3$} & \multicolumn{4}{c}{$d_m=4$} & \multicolumn{4}{c}{$d_m=5$} \\
    \cmidrule(lr){2-5} \cmidrule(lr){6-9} \cmidrule(lr){10-13} \cmidrule(lr){14-17} \cmidrule(lr){18-21}
     & 0 & 1 & 2 & M & 0 & 1 & 2 & M & 0 & 1 & 2 & M & 0 & 1 & 2 & M & 0 & 1 & 2 & M \\
    \hline
    Proj-GAN &  0.040 &  0.106 &  0.273 &  0.074 &  0.044 &  0.327 &  1.246 &  0.248 &  0.060 &  0.635 &  1.628 &  0.325 &  0.073 &  0.932 &  3.379 &  0.527 &  0.166 &  3.298 &  3.903 &  1.126 \\
    TAC-GAN* &  {\bf 0.015 } &  {\bf 0.033 } &  {\bf 0.100 } &  {\bf 0.027 } &  0.021 &  0.124 &  0.529 &  0.067 &  {\bf 0.020 } &  0.272 &  0.803 &  0.149 &  {\bf 0.027 } &  {\bf 0.412 } &  {\bf 1.969 } &  {\bf 0.106 } &  {\bf 0.035 } &  1.139 &  {\bf 2.160 } &  {\bf 0.156 } \\
    $f$-cGAN &  0.018 &  0.042 &  0.170 &  0.031 &  {\bf 0.019 } &  {\bf 0.090 } &  {\bf 0.383 } &  {\bf 0.047 } &  0.030 &  {\bf 0.193 } &  {\bf 0.635 } &  {\bf 0.087 } &  {\bf 0.024 } &  0.575 &  2.170 &  0.276 &  0.037 &  {\bf 0.857 } &  3.328 &  0.287 \\
    P2GAN    &  {\bf 0.009 } &  {\bf 0.028 } &  {\bf 0.151 } &  {\bf 0.026 } &  {\bf 0.014 } &  {\bf 0.080 } &  {\bf 0.345 } &  {\bf 0.046 } &  {\bf 0.016 } &  {\bf 0.160 } &  {\bf 0.639 } &  {\bf 0.084 } &  0.028 &  {\bf 0.237 } &  {\bf 1.530 } &  {\bf 0.056 } &  {\bf 0.030 } &  {\bf 0.655 } &  {\bf 2.725 } &  {\bf 0.261 } \\
    \hline
    Proj-GAN &  {\bf 0.112 } &  {\bf 0.267 } &  {\bf 0.879 } &  {\bf 0.178 } &  {\bf 0.167 } &  {\bf 0.725 } &  {\bf 2.373 } &  {\bf 0.492 } &  {\bf 0.172 } &  {\bf 1.455 } &  {\bf 6.385 } &  {\bf 0.969 } &  0.249 &  {\bf 4.904 } &  {\bf 15.496 } &  {\bf 3.368 } &  0.386 &  11.002 &  {\bf 29.382 } &  {\bf 7.407 } \\
    TAC-GAN* &  0.190 &  0.474 &  {\bf 1.376 } &  {\bf 0.416 } &  0.304 &  1.635 &  {\bf 4.213 } &  1.060 &  0.357 &  3.504 &  {\bf 12.817 } &  {\bf 2.531 } &  0.314 &  6.949 &  29.822 &  6.425 &  {\bf 0.264 } &  13.905 &  54.134 &  11.125 \\
    $f$-cGAN &  0.164 &  {\bf 0.429 } &  1.484 &  0.441 &  0.192 &  1.718 &  5.084 &  1.506 &  0.174 &  3.480 &  15.491 &  3.675 &  {\bf 0.138 } &  {\bf 3.629 } &  {\bf 19.597 } &  {\bf 3.862 } &  {\bf 0.173 } &  {\bf 4.592 } &  {\bf 28.753 } &  {\bf 4.315 } \\
    P2GAN    &  {\bf 0.118 } &  0.584 &  1.696 &  0.490 &  {\bf 0.120 } &  {\bf 0.843 } &  4.486 &  {\bf 1.051 } &  {\bf 0.152 } &  {\bf 2.951 } &  12.854 &  2.852 &  {\bf 0.192 } &  6.005 &  22.003 &  5.066 &  0.295 &  {\bf 9.920 } &  36.080 &  8.145 \\
    \bottomrule
    \end{tabular}}
\end{table*}
\begin{figure*}[h!]
  \begin{center}
    \subfloat[Proj-GAN]{\includegraphics[width=0.32\linewidth]{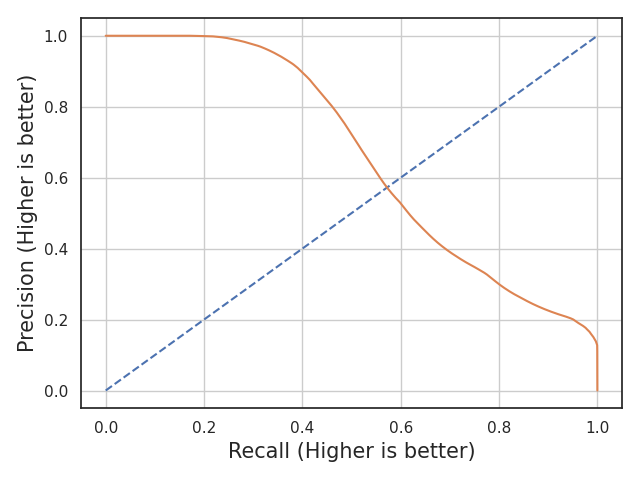}}\hfill
    \subfloat[P2GAN-w]{\includegraphics[width=0.32\linewidth]{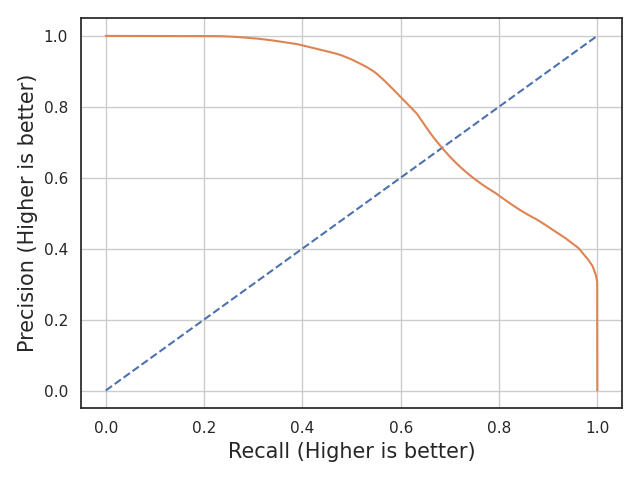}}\hfill
    \subfloat[P2GAN]{\includegraphics[width=0.32\linewidth]{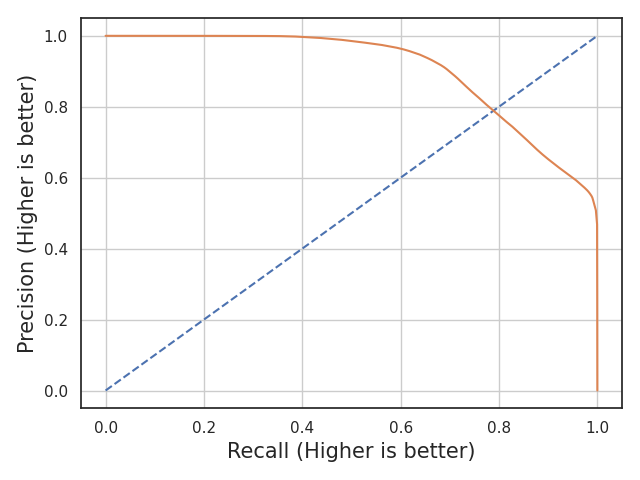}}\hfill
    \caption{Precision \& recall on ImageNet (128 resolution).}
    \label{fig:pr}
  \end{center}
\end{figure*}

\section{CIFAR}
\noindent \textbf{Experimental setup.} To construct the CIFAR100IB dataset, we randomly sample $N_c$ images from class $c$ where $N_c = \text{round}(500-4\times c)$. For CIFAR100 experiments, we fix batch size as 100, and the number of $D$ steps per $G$ step as 4. All baselines are trained for 500 epochs or 62k iterations. These hyper-parameters are kept the same as described in TAC-GAN paper (also in their provided launch script). 

\begin{figure*}[t]
  \begin{center}
    \hfill
    \subfloat[P2GAN]{\includegraphics[width=0.44\linewidth]{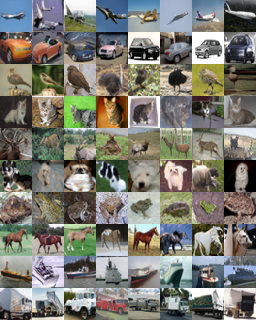}}\hfill
    \subfloat[P2GAN-w]{\includegraphics[width=0.44\linewidth]{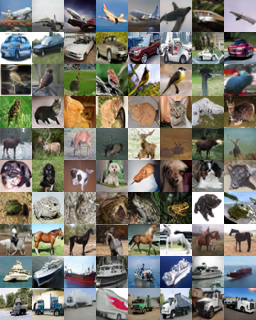}}\hfill
    \hfill
    \caption{10 classes of CIFAR10 generated samples at $32 \times 32$ resolution.}
    \label{fig:sample_c10}
  \end{center}
\end{figure*}

\begin{figure*}[t]
  \begin{center}
    \subfloat[P2GAN]{\includegraphics[width=0.95\linewidth]{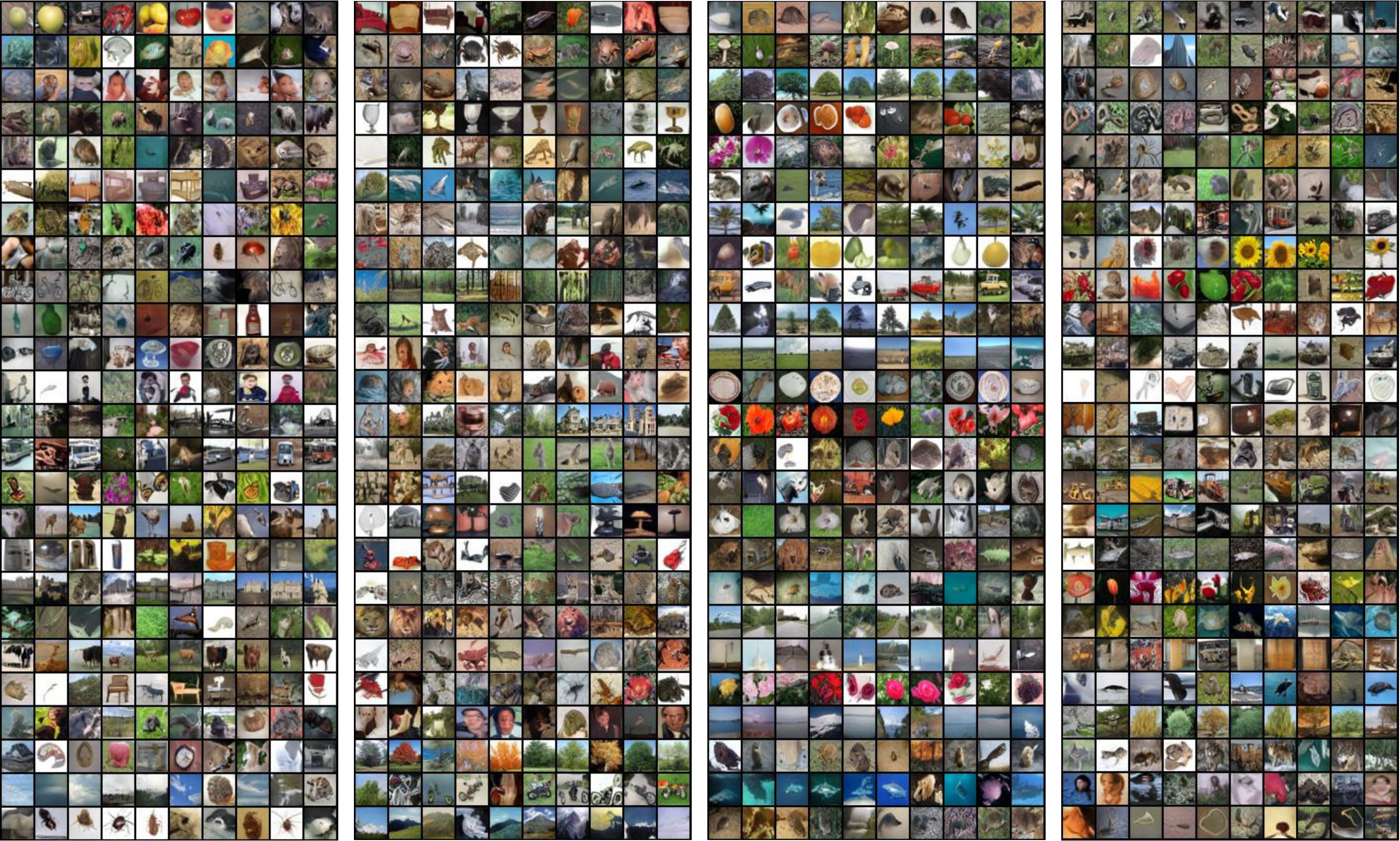}}\par
    \subfloat[P2GAN-w]{\includegraphics[width=0.95\linewidth]{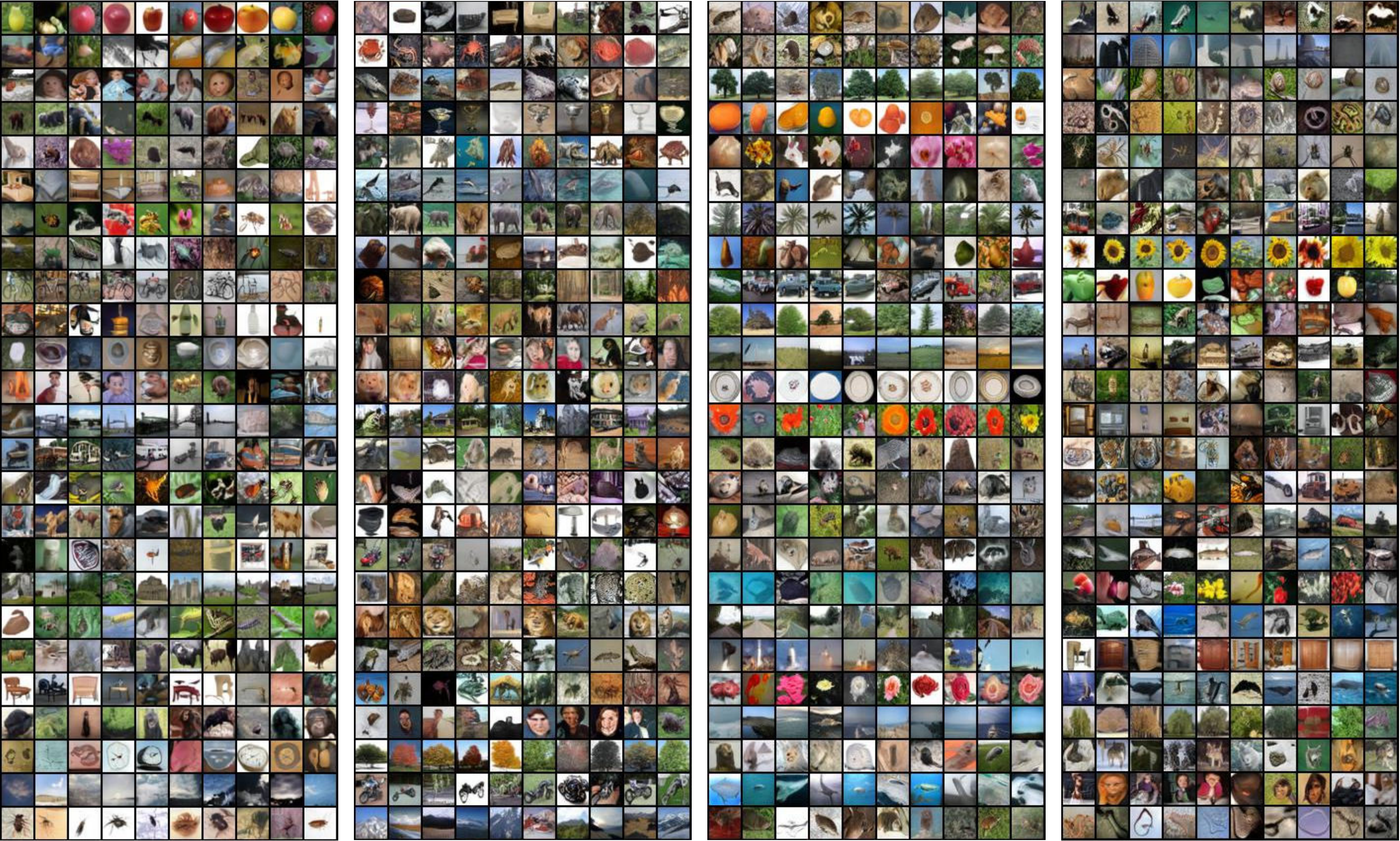}}
    \caption{100 classes of CIFAR100 generated samples at $32 \times 32$ resolution.}
    \label{fig:sample_c100}
  \end{center}
\end{figure*}

\noindent \textbf{More results.} Generated samples of CIFAR10 and CIFAR100 are shown in Figure~\ref{fig:sample_c10} and~\ref{fig:sample_c100}, respectively. 

\section{ImageNet}
\noindent \textbf{Experimental setup.} Due to limited computation resource, experiments on ImageNet related to model comparison and analysis are conducted at resolution $64 \times 64$. We follow the experimental setup in TAC-GAN paper but reduce the image size and model size. We use batch size of 2048 (batch size 256 accumulated 8 times) and the number of $D$ steps per $G$ step is $1$. Channel multipliers for both $G$ and $D$ are $32$. The resolution of self-attention layer is set to $32$. Models are trained with $80\text{k}$ iterations.

For experiments at $128 \times 128$ resolution, we follow the configurations of BigGAN256\footnote{\url{https://github.com/POSTECH-CVLab/PyTorch-StudioGAN/blob/master/src/configs/ILSVRC2012/BigGAN256.json}} provided in the StudioGAN~\cite{kang2020contragan} repo. We use batch size of 256 and the number of $D$ steps per $G$ step is $2$. Channel multipliers for both $G$ and $D$ are $96$. The resolution of self-attention layer is set to $64$. We train a P2GAN-w model with $200\text{k}$ iterations.

\noindent \textbf{More results.}
Although P2GAN (without adaptive weights) achieves the highest IS, it shows {\it mode collapse} on certain classes (for example the ``flowers'' in row 45). For the same flower class, Proj-GAN can still generate diverse samples. TAC-GAN, $f\mhyphen\text{cGAN}$ and P2GAN all exhibit mode collapse on certain classes. While the proposed weighting strategy is able to avoid mode collapse and still achieve competitively high IS and low FID.

\begin{figure*}[h]
    \centering
    \includegraphics[width=0.95\linewidth]{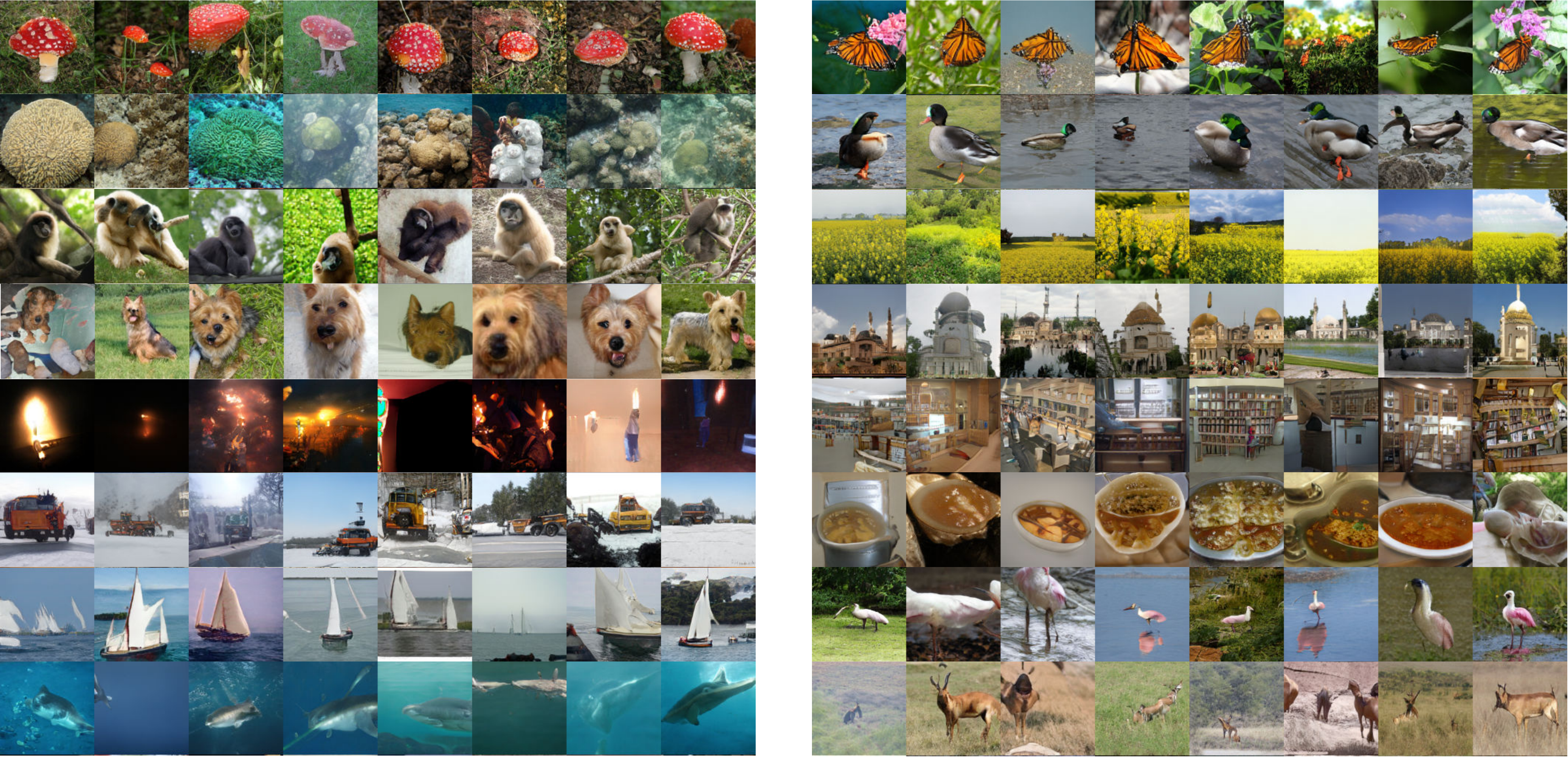}
    \caption{P2GAN-w generated samples on ImageNet at $128 \times 128$ resolution.}
    \label{fig:sample_i128}
\end{figure*}

Results of $128 \times 128$ resolution ImageNet experiments are reported in Figure~\ref{fig:train_i128}. The IS and FID curves over training iterations clearly shows its advantage in terms of fast convergence. Precision-recall~\cite{sajjadi2018assessing} curves are given in Figure~\ref{fig:pr}. Some randomly generated samples of P2GAN-w during training are shown in Figure~\ref{fig:sample_i128}.

\section{VGGFace2}
\noindent \textbf{Experimental setup.}
We follow the same protocol as in TAC-GAN paper, and set batch size to 256 and the number of $D$ steps per $G$ step to 1. Images are resized to resolution of $64 \times 64$. The resolution of self-attention layer is set to 32. Channel multipliers for both $G$ and $D$ are $32$. All baselines are trained with 100k iterations.

As for evaluation, we tried our best effort to match the calculated FID and IS with the reported values in TAC-GAN~\cite{gong2019twin}. However, these values can be affected by many factors such as the selected subset of identities and the checkpoint of Inception Net~\cite{szegedy2016rethinking} used for evaluation. We first sample a subset of 2000 identities and finetune an Inception model using Adam optimizer~\cite{kingma2014adam}. We use the checkpoinit at 20000 iteration to monitor the training of GAN models. Then we train a TAC-GAN model\footnote{Here the model is chosen to be its actual implementation, which is equivalent to $f\mhyphen\text{cGAN}$ with reverse-KL and cross-entropy loss.} and select the best model with the lowest FID. Finally, we use the selected TAC-GAN model to examine which Inception Net checkpoint yields the best match. The final FID score is 29.54 which is very close to the reported 29.12. The identities of subsets VGGFace200, VGGFace500 and VGGFace2000 are given in Supplemental Materials.

\begin{figure*}[h]
    \centering
    \includegraphics[width=0.9\linewidth]{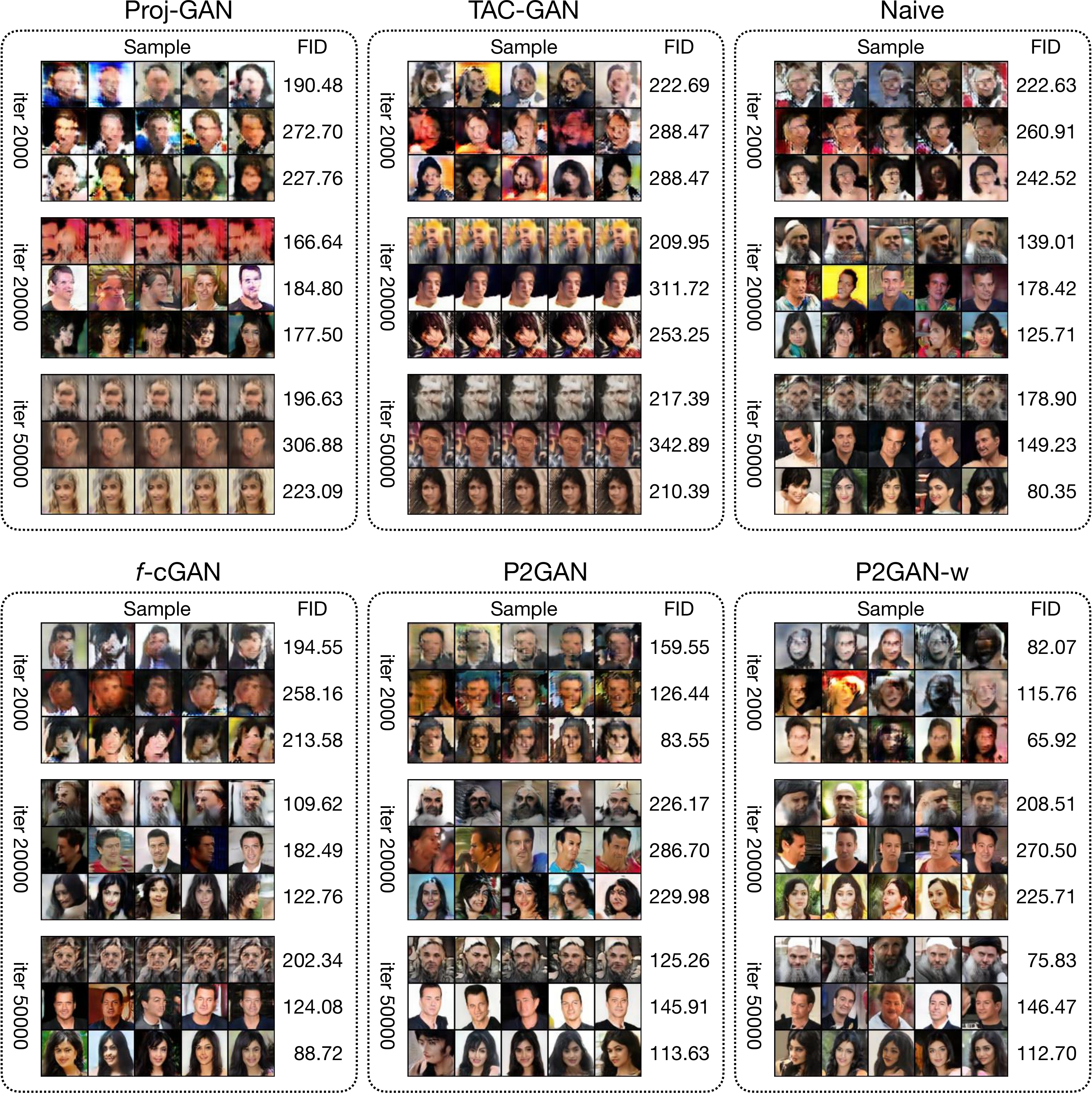}
    \caption{Samples and FID scores of VGGFace200, evaluated at iteration 2000, 20000, and 50000. Their identity numbers are 0, 1, and 2, respectively. At iteration 50000, all methods except for P2GAN-w exhibit mode collapse on identity 0.}
    \label{fig:collapse_v200}
\end{figure*}

\noindent \textbf{More results.}
As a complementary to the t-SNE visualization of image embeddings provided in the main text, we visualize the samples and list the corresponding FID values in Figure~\ref{fig:collapse_v200}. We see that Proj-GAN, TAC-GAN, $f\mhyphen\text{cGAN}$ and P2GAN all show {\it mode collapse} on identity 0 (the first row) while P2GAN-w still generates diverse samples on the given class. Additional IS and FID values are reported in Table~\ref{tab:collapse_v200}.

The training curves of different baselines on VGGFace200 are plotted in Figure~\ref{fig:train_v200}. We see that Proj-GAN, over-parameterzation baseline ($\lambda \equiv 0$), DM-GAN ($\psi \equiv 0$) and the na\"ive baseline all fail on VGGFace200. 


\end{document}